\documentclass{article}

   \PassOptionsToPackage{numbers, compress}{natbib}

    \usepackage[preprint]{neurips_2020}

\usepackage[utf8]{inputenc} 
\usepackage[T1]{fontenc}    
\usepackage{hyperref}       
\usepackage{url}            
\usepackage{booktabs}       
\usepackage{amsfonts}       
\usepackage{nicefrac}       
\usepackage{microtype}      
\usepackage{wrapfig}
\usepackage{graphicx}
\usepackage{amssymb}
\usepackage{amsmath}
\usepackage{mathrsfs}
\usepackage{multirow}
\usepackage{algorithm}
\usepackage{algorithmic}
\usepackage{subfigure}
\usepackage{booktabs}
\usepackage{enumitem}
\usepackage{caption}

\newtheorem{theorem}{Theorem}
\newtheorem{lemma}[theorem]{Lemma}

\newtheorem{definition}[theorem]{Definition}
\newtheorem{proposition}[theorem]{Proposition}

\newtheorem{corollary}[theorem]{Corollary}

\title{Dynamic of Stochastic Gradient Descent with State-Dependent Noise}

\author{
  Qi Meng$^1$\thanks{Corresponding E-mails: meq@microsoft.com, wche@microsoft.com}, Shiqi Gong$^2$, Wei Chen$^1$, Zhi-Ming Ma$^3$, Tie-Yan Liu$^1$\\
  $^1$Microsoft Research Asia, $^{2,3}$University of Chinese Academy of Sciences\\
  $^1\{$meq, wche, tie-yan.liu$\}$@microsoft.com,\\
   $^2$gongshiqi15@mails.ucas.ac.cn, $^3$mazm@amt.ac.cn
}

\begin{document}

\maketitle

\begin{abstract}
Stochastic gradient descent (SGD) and its variants are mainstream methods to train deep neural networks. Since neural networks are non-convex, more and more works study the dynamic behavior of SGD and its impact to generalization, especially the escaping efficiency from local minima. However, these works make the over-simplified assumption that the distribution of gradient noise is state-independent, although it is state-dependent. In this work, we propose a novel power-law dynamic with state-dependent diffusion to approximate the dynamic of SGD. Then, we prove that the stationary distribution of power-law dynamic is heavy-tailed, which matches the existing empirical observations. Next, we study the escaping efficiency from local minimum of power-law dynamic and prove that the mean escaping time is in polynomial order of the barrier height of the basin, much faster than exponential order of previous dynamics. It indicates that SGD can escape deep sharp minima efficiently and tends to stop at flat minima that have lower generalization error. 
Finally, we conduct experiments to compare SGD and power-law dynamic, and the results verify our theoretical findings.
\end{abstract}

	\section{Introduction}
Deep learning has achieved great success in various AI applications, such as computer vision, natural language processing, and speech recognition \cite{he2016deep,vaswani2017attention,he2016dual}. Stochastic gradient descent (SGD) and its variants are the mainstream methods to train deep neural networks, since they can deal with the computational bottleneck of the training over large-scale datasets \cite{bottou2008tradeoffs}. 

Although SGD can converge to the minimum in convex optimization \cite{rakhlin2012making}, neural networks are highly non-convex. To understand the behavior of SGD on non-convex optimization landscape, on one hand, researchers are investigating the loss surface of the neural networks with variant architectures \cite{choromanska2015loss,li2018visualizing,he2019asymmetric,draxler2018essentially,li2018over}; on the other hand, researchers illustrate that the noise in stochastic algorithm may make it escape from local minima \cite{keskar2016large,he2019control,zhu2019anisotropic,wu2019multiplicative, haochen2020shape}.  
It is clear that whether stochastic algorithms can escape from poor local minima and finally stop at a minimum with low generalization error is crucial to its test performance.
In this work, we focus on the dynamic of SGD and its impact to generalization, especially the escaping efficiency from local minima. 

To study the dynamic behavior of SGD, most of the works consider SGD as the discretization of a continuous-time dynamic system and investigate its dynamic properties.
There are two typical types of models to approximate dynamic of SGD.  \cite{li2017stochastic, zhou2019toward, liu2018toward, chaudhari2018stochastic, he2019control, zhu2019anisotropic,hu2019diffusion,xie2020diffusion} approximate the dynamic of SGD by Langevin dynamic with constant diffusion coefficient and proved its escaping efficiency from local minima.
These works make over-simplified assumption that the covariance matrix of gradient noise is constant, although it is state-dependent in general. The simplified assumption makes the proposed dynamic unable to explain the empirical observation that the distribution of parameters trained by SGD is heavy-tailed \cite{mahoney2019traditional}. To model the heavy-tailed phenomenon,  \cite{simsekli2019tail,csimcsekli2019heavy} point that the variance of stochastic gradient may be infinite, and they propose to approximate SGD by dynamic driven by $\alpha$-stable process 
with the strong infinite variance condition. However, as shown in the work \cite{xie2020diffusion, mandt2017stochastic}, the gradient noise follows Gaussian distribution and the infinite variance condition does not satisfied.  Therefore it is still lack of suitable theoretical explanation on the implicit regularization of dynamic of SGD. 

In this work, we conduct a formal study on the (state-dependent) noise structure of SGD and its dynamic behavior. First, we show that the covariance of the noise of SGD in the quadratic basin surrounding the local minima is a quadratic function of the state (i.e., the model parameter). Thus, we propose approximating the dynamic of SGD near the local minimum using a stochastic differential equation whose diffusion coefficient is a quadratic function of state. We call the new dynamic \textit{power-law dynamic}. We prove that its stationary distribution is power-law $\kappa$ distribution, 
where $\kappa$ is the signal to noise ratio of the second order derivatives at local minimum. Compared with Gaussian distribution, power-law $\kappa$ distribution is heavy-tailed with tail-index $\kappa$.  It matches the empirical observation that the distribution of parameters becomes heavy-tailed after SGD training without assuming infinite variance of stochastic gradient in \cite{simsekli2019tail}. 

Second, we analyze the escaping efficiency of power-law dynamic from local minima and its relation to generalization.  By using the random perturbation theory for diffused dynamic systems, we analyze the mean escaping time for power-law dynamic. 
Our results show that: (1) Power-law dynamic can escape from sharp minima faster than flat minima. (2) The mean escaping time for power-law dynamic is only in the polynomial order of the barrier height, much faster than the exponential order for dynamic with constant diffusion coefficient. 
Furthermore, we provide a PAC-Bayes generalization bound and show power-law dynamic can generalize better than dynamic with constant diffusion coefficient. Therefore, our results indicate that the state-dependent noise helps SGD to escape from sharp minima quickly and implicitly learn well-generalized model.

Finally, we corroborate our theory by experiments. We investigate the distributions of parameters trained by SGD on various types of deep neural networks and show that they are well fitted by power-law $\kappa$ distribution. Then, we compare the escaping efficiency of dynamics with constant diffusion or state-dependent diffusion to that of SGD. Results show that the behavior of power-law dynamic is more consistent with SGD. 

Our contributions are summarized as follows: (1)  We propose a novel power-law dynamic with state-dependent diffusion to approximate dynamic of SGD based on both theoretical derivation and empirical evidence. The power-law dynamic can explain the heavy-tailed phenomenon of parameters trained by SGD without assuming infinite variance of gradient noise.  (2) We analyze the mean escaping time and PAC-Bayes generalization bound for power-law dynamic and results show that power-law dynamic can escape sharp local minima faster and generalize better compared with the dynamics with constant diffusion. Our experimental results can support the theoretical findings.


\section{Background}
\label{sec:background}
In empirical risk minimization problem, the objective is 
$L(w)=\frac{1}{n}\sum_{i=1}^n\ell(x_i,w)$, where $x_i,i=1,\cdots,n$ are $n$ \textit{i.i.d.} training samples,  $w\in\mathbb{R}^d$ is the model parameter, and $\ell$ is the loss function. Stochastic gradient descent (SGD) is a popular optimization algorithm to minimize $L(w)$. The update rule is 
$w_{t+1}=w_t-\eta\cdot \tilde{g}(w_t),$ where $\tilde{g}(w_t)=\frac{1}{b}\sum_{x\in S_b}\nabla_w\ell(x,w_t)$ is the minibatch gradient calculated by a randomly sampled minibatch $S_b$ of size $b$ and $\eta$ is the learning rate.  The minibatch gradient $\tilde{g}(w_t)$ is an unbiased estimator of the full gradient $g(w_t)=\nabla L(w_t)$, and
the term $ (g(w_t)-\tilde{g}(w_t))$ is called \emph{gradient noise} in SGD.

\textbf{Langevin Dynamic}
In \cite{li2017stochastic,he2019control}, the gradient noise is assumed to be drawn from Gaussian distribution according to central limit theorem (CLT), 
i.e.,
$g(w)-\tilde{g}(w)\sim\mathcal{N}(0,C),$ where covariance matrix  $C$ is a constant matrix for all $w$. 
Then SGD can be regarded as the numerical discretization of the following Langevin dynamic,
\begin{equation}\label{eq2.2}
dw_t=-g(w_t) dt+\sqrt{\eta}C^{1/2}dB_t,
\end{equation}
where $B_t$ is a standard Brownian motion in $\mathbb{R}^d$ and $\sqrt{\eta}C^{1/2}dB_t$ is called the diffusion term.



\textbf{$\alpha$-stable Process}  \cite{simsekli2019tail} assume the variance of gradient noise is unbounded. By generalized CLT, the distribution of gradient noise is $\alpha$-stable distribution $\mathcal{S}(\alpha,\sigma)$, where $\sigma$ is the $\alpha$-th moment of gradient noise for given $\alpha$ with $\alpha\in(0,2]$. Under this assumption, SGD is approximated by the stochastic differential equation (SDE) driven by an $\alpha$-stable process.


\subsection{Related Work}
There are many works that approximate SGD by Langevin dynamic with constant diffusion coefficient. From the aspect of optimization,  the convergence rate of SGD and its optimal hyper-parameters have been studied in \cite{li2017stochastic,he2018differential,liu2018toward,he2018differential} via optimal control theory.
From the aspect of generalization, \cite{chaudhari2018stochastic,zhang2018energy,smith2017bayesian} show that SGD implicitly regularizes the negative entropy of the learned distribution. Recently, the escaping efficiency from local minima of Langevin dynamic has been studied \cite{zhu2019anisotropic,hu2019diffusion,xie2020diffusion}. \cite{he2019control} analyze the PAC-Bayes generalization error of Langevin dynamic to explain the generalization of SGD.  

The solution of Langevin dynamic is Gaussian distribution, which does not match the empirical observations that the distribution of parameters trained by SGD is a heavy-tailed \cite{mahoney2019traditional,hodgkinson2020multiplicative,gurbuzbalaban2020heavy}. 
\cite{simsekli2019tail,csimcsekli2019heavy} assume the variance of stochastic gradient is infinite and regard SGD as discretization of a stochastic differential equation (SDE) driven by an $\alpha$-stable process. The escaping efficiency for the SDE is also shown in \cite{simsekli2019tail}. 

However, these theoretical results are derived for dynamics with constant diffusion term, although the gradient noise in SGD is state-dependent.  
There are some related works analyze state-dependent noise structure in SGD, such as label noise in \cite{haochen2020shape} and multiplicative noise in \cite{wu2019noisy}.
These works propose new algorithms motivated by the noise structure, but they do not analyze the escaping behavior of dynamic of SGD and the impact to generalization.

\section{Approximating SGD by Power-law Dynamic}
In this section, we study the (state-dependent) noise structure of SGD (in Section 3.1) and propose power-law dynamic to approximate the dynamic of SGD. We first study 1-dimensional power-law dynamic in Section 3.2 and extend it to high dimensional case in Section 3.3.

\subsection{Noise Structure of Stochastic Gradient Descent}
For non-convex optimization, we investigate the noise structure of SGD around local minima so that we can analyze the escaping efficiency from it. We first describe the quadratic basin where the local minimum is located. 
Suppose $w^*$ is a local minimum of the training loss $L(w)$ and $g(w^*)=0$.
We name the $\epsilon$-ball $\mathcal{B}(w^*,\epsilon)$ with center $w^*$ and radius $\epsilon$ as a quadratic basin if the loss function for $w\in\mathcal{B}(w^*,\epsilon)$ can be approximated by second-order Taylor expansion as $L(w)= L(w^*)+\frac{1}{2} (w-w^*)^T H(w^*)(w-w^*)$. Here, $H(w^*)$ is the Hessian matrix of loss at $w^*$, which is (semi) positive definite.

Then we start to analyze the gradient noise of SGD. The full gradient of training loss is $g(w)=H(w^*)(w-w^*)$. The stochastic gradient is $\tilde{g}(w)=\tilde{g}(w^*)+\tilde{H}(w^*)(w-w^*)$ by Taylor expansion where $\tilde{g}(\cdot)$ and $\tilde{H}(\cdot)$ are stochastic version of gradient and Hessian calculated by the minibatch.
The randomness of gradient noise comes from two parts:  $\tilde{g}(w^*)$ and $\widetilde{H}(w^*)$, which reflects the fluctuations of the first-order and second-order derivatives of the model at $w^*$ over different minibatches, respectively. The following proposition gives the variance of the gradient noise. 
\begin{proposition}\label{propo1}
	For $w\in\mathcal{B}(w^*,\epsilon)\subset\mathbb{R}$, the variance of gradient noise is 
	$\sigma(g(w)-\tilde{g}(w))=\sigma(\tilde{g}(w^*))+2\rho(\tilde{g}(w^*),\tilde{H}(w^*))(w-w^*)+\sigma(\tilde{H}(w^*))(w-w^*)^2$, where $\sigma(\cdot)$ and $\rho(\cdot,\cdot)$ are the variance and covariance in terms of the minibatch.
\end{proposition}
From Proposition \ref{propo1}, we can conclude that: (1) The variance of noise is finite if $\tilde{g}(w^*)$ and $\tilde{H}(w^*)$ have finite variance because {\small${\rho(\tilde{g}(w^*),\tilde{H}(w^*))\leq\sqrt{\sigma(\tilde{g}(w^*))\cdot\sigma(\tilde{H}(w^*))}}$ } according to Cauchy–Schwarz inequality. For fixed $w^*$, a sufficient condition for that $\tilde{g}(w^*)$ and $\tilde{H}(w^*)$ have finite variance is that the training data $x$ are sampled from bounded domain. This condition is easy to be satisfied because the domain of training data are usually normalized to be bounded before training. In this case, the infinite variance assumption about the stochastic gradient in $\alpha$-stable process is satisfied. 
(2) The variance of noise is state-dependent, which contradicts the assumption in Langevin dynamic. 

\textbf{Notations:} For ease of the presentation, we use {\small$C(w), \sigma_g, \sigma_H, \rho_{g,H}$} to denote {\small$\sigma(g(w)-\tilde{g}(w^*))$, $\sigma(\tilde{g}(w^*))$, $\sigma(\tilde{H}(w^*))$, $\rho(\tilde{g}(w^*),\tilde{H}(w^*))$} in the following context, respectively.
\subsection{Power-law Dynamic}  
According to CLT, the gradient noise follows Gaussian distribution if it has finite variance, i.e.,
\begin{align}
g(w)-\tilde{g}(w)\rightarrow_p \mathcal{N}(0, C(w))\quad \textit{as}\quad b\rightarrow\infty,
\end{align}where $\rightarrow_p$ means “converge in distribution”. Using Gaussian distribution to model the gradient noise in SGD, the update rule of SGD can be written as:
\begin{align}
w_{t+1}=w_t-\eta g(w_t)+ \eta \xi_t, \quad \xi_t\sim\mathcal{N}(0,C(w)).
\end{align}
Eq.\ref{pld} can be treated as the discretization of the following SDE, which we call it power-law dynamic:
\begin{align}\label{pld}
dw_t=-g(w_t)dt+\sqrt{\eta C(w)}dB_t.
\end{align}
Power-law dynamic characterizes how the distribution of $w$ changes as time goes on. 
The distribution density of parameter $w$ at time $t$ (i.e., $p(w,t)$) is determined by the Fokker-Planck equation (Zwanzig's type \cite{guo2014power}):
\begin{equation}
\frac{\partial}{\partial t}p(w,t)=\nabla p(w,t)g(w)+\frac{\eta}{2}\cdot\nabla\left(C(w)\cdot\nabla p(w,t)\right).
\end{equation}
The stationary distribution of power-law dynamic can be obtained if we let the left side of Fokker-Planck equation be zero. 
The following theorem shows the analytic form of the stationary distribution of power-law dynamic, which is heavy-tailed and the tail of the distribution density decays at polynomial order of $w-w^*$.  This is the reason why we call the stochastic differential equation in Eq.\ref{pld} power-law dynamic.    
\begin{theorem}
	The stationary distribution density for 1-dimensional power-law dynamic (Eq.\ref{pld}) is
	{\small\begin{align}\label{eq:densfull}
		p(w)=\frac{1}{Z}(C(w))^{-\frac{H}{\eta\sigma_{{H}}}}\exp\left(-
		\frac{H\left( 4\rho_{g,H}\cdot ArcTan\left(C'(w)/\sqrt{4\sigma_H\sigma_g-4\rho_{g,H}}\right)\right)}{\eta\sigma_H}\right),
		\end{align}}where $Z$ is the normalization constant to make $\int_{-\infty}^{\infty}p(w)dw=1$ and {$ArcTan(\cdot)$} is the arctangent function. Furthermore, $p(w)$ is heavy-tailed, i.e., {$\lim_{w\rightarrow\infty}e^{cw}p(w)=\infty$} for all $c>0$.
\end{theorem}
We explain more about the heavy-tailed property of $p(w)$. Because the function {\small$ArcTan(\cdot)$} is bounded, the decreasing rate of $p(w)$ as $w\rightarrow\infty$ is mainly determined by the term {\small$C(w)^{-\frac{H}{\eta\sigma_H}}$}, which is a polynomial function about $w-w^*$. Therefore, the decreasing rate of $p(w)$ is slower than $e^{tw}$. Here, we call $\frac{H}{\eta\sigma_H}$ the \emph{tail-index} of $p(w)$ and denote it as $\kappa$ in the following context. 

We can conclude that the state-dependent noise results in heavy-tailed distribution of parameters, which matches the observations in \cite{mahoney2019traditional}. 
Langevin dynamic with constant diffusion can be regarded as special case of power-law dynamic when $\rho_{H,g}=0$ and $\sigma_H=0$. In this case, $p(w)$ degenerates to Gaussian distribution. Compared with $\alpha$-stable process, we do not assume infinite variance on gradient noise and demonstrate another mechanism that results in heavy-tailed distribution of parameters.                         \begin{figure}
	\centering
	\subfigure[CNN layer-1]{\label{fig:qua2}\includegraphics[width=0.20\textwidth]{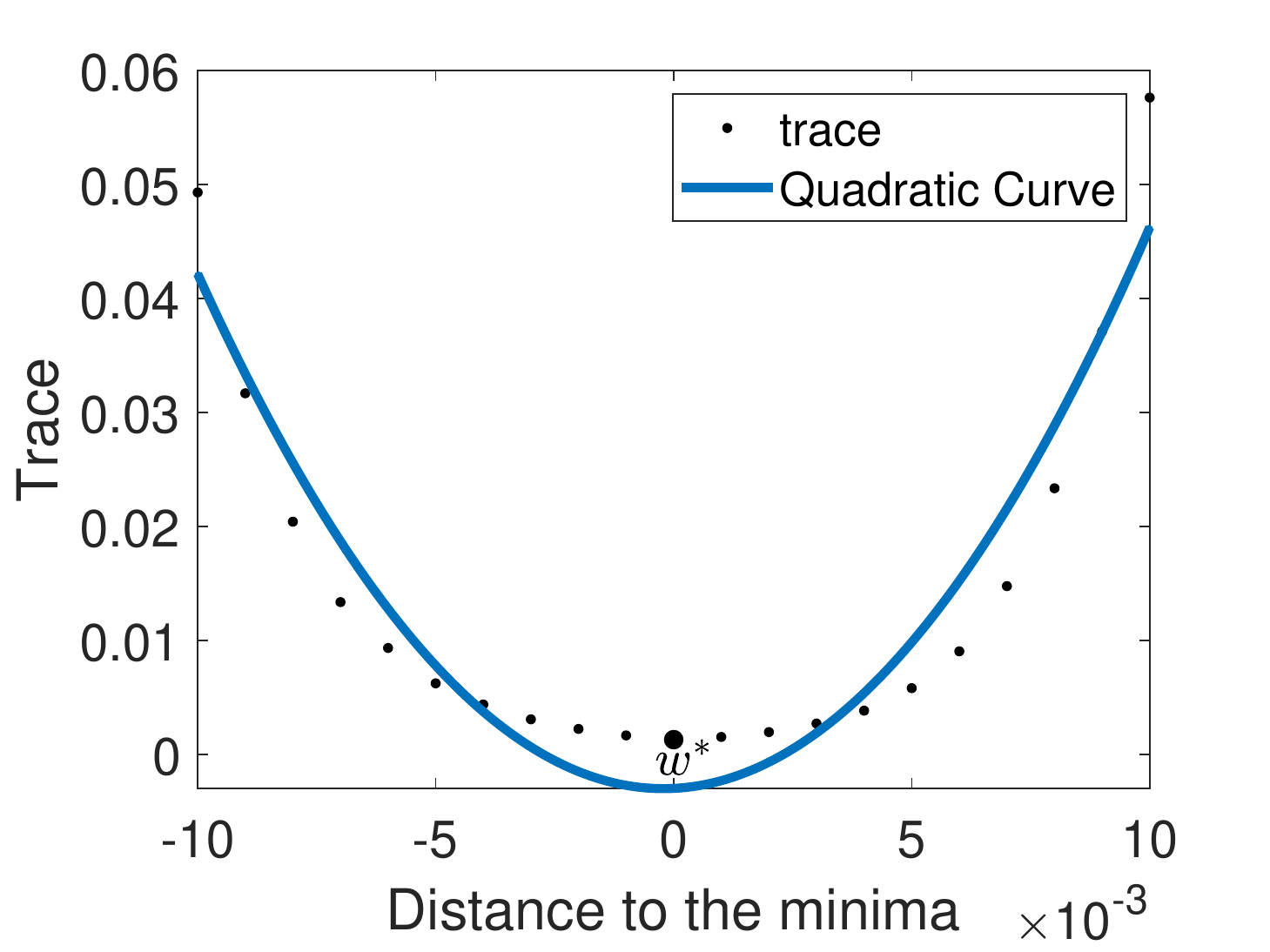}}
	\subfigure[CNN layer-2]{\label{fig:qua}\includegraphics[width=0.20\textwidth]{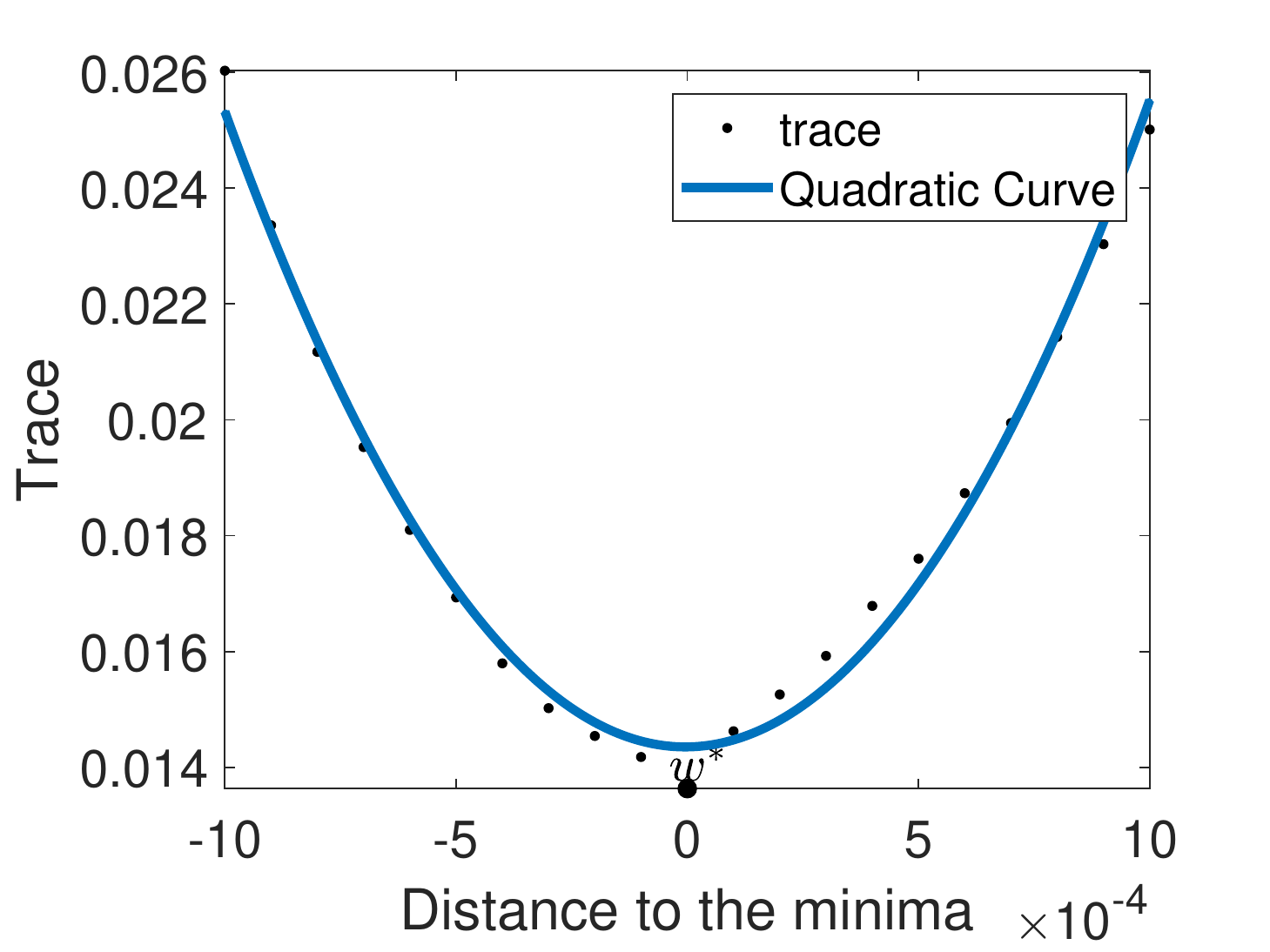}}
	\subfigure[ResNet layer-1]{\label{fig:qua3}\includegraphics[width=0.20\textwidth]{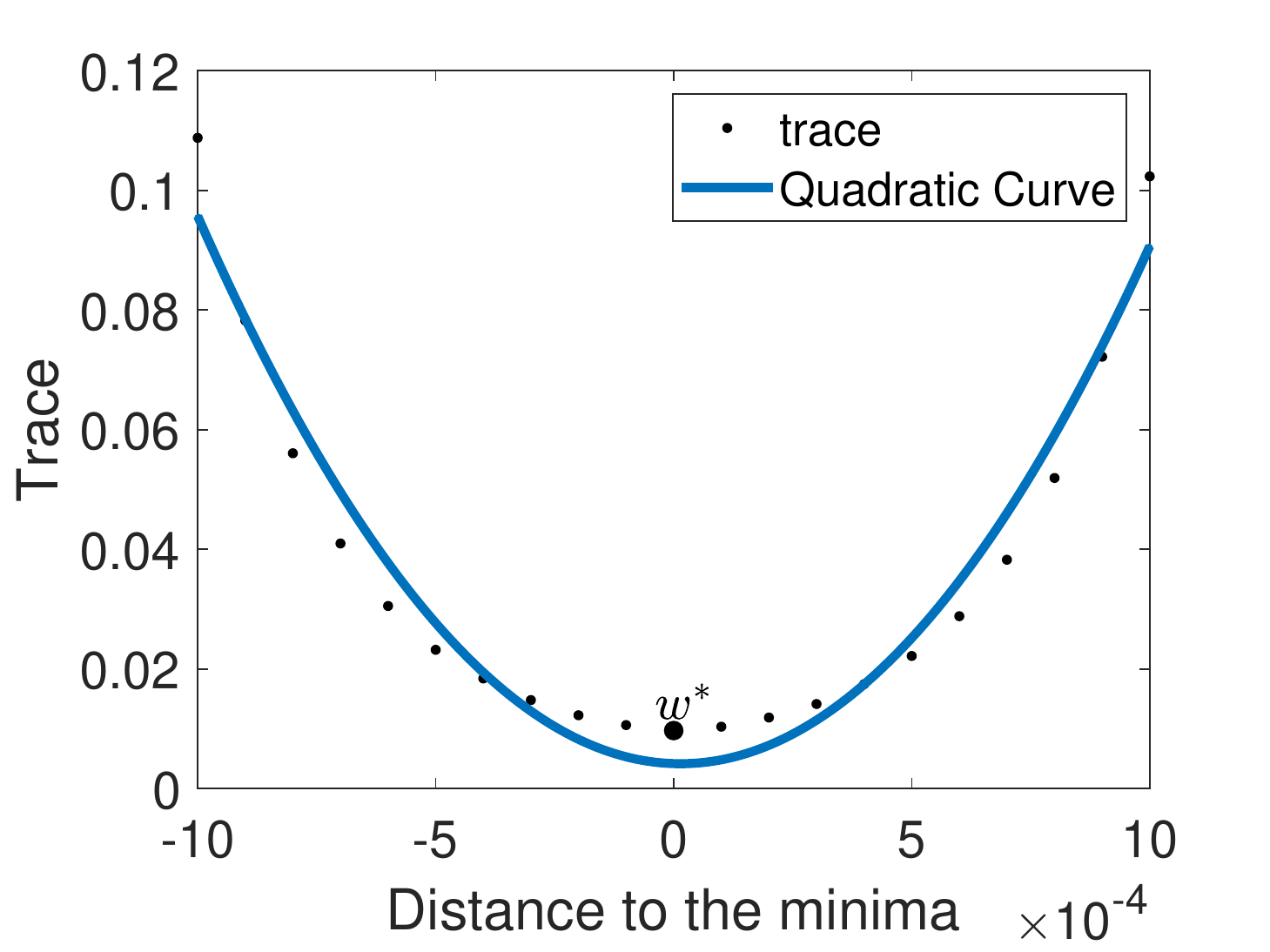}}
	\subfigure[ResNet layer-2]{\label{fig:qua4}\includegraphics[width=0.20\textwidth]{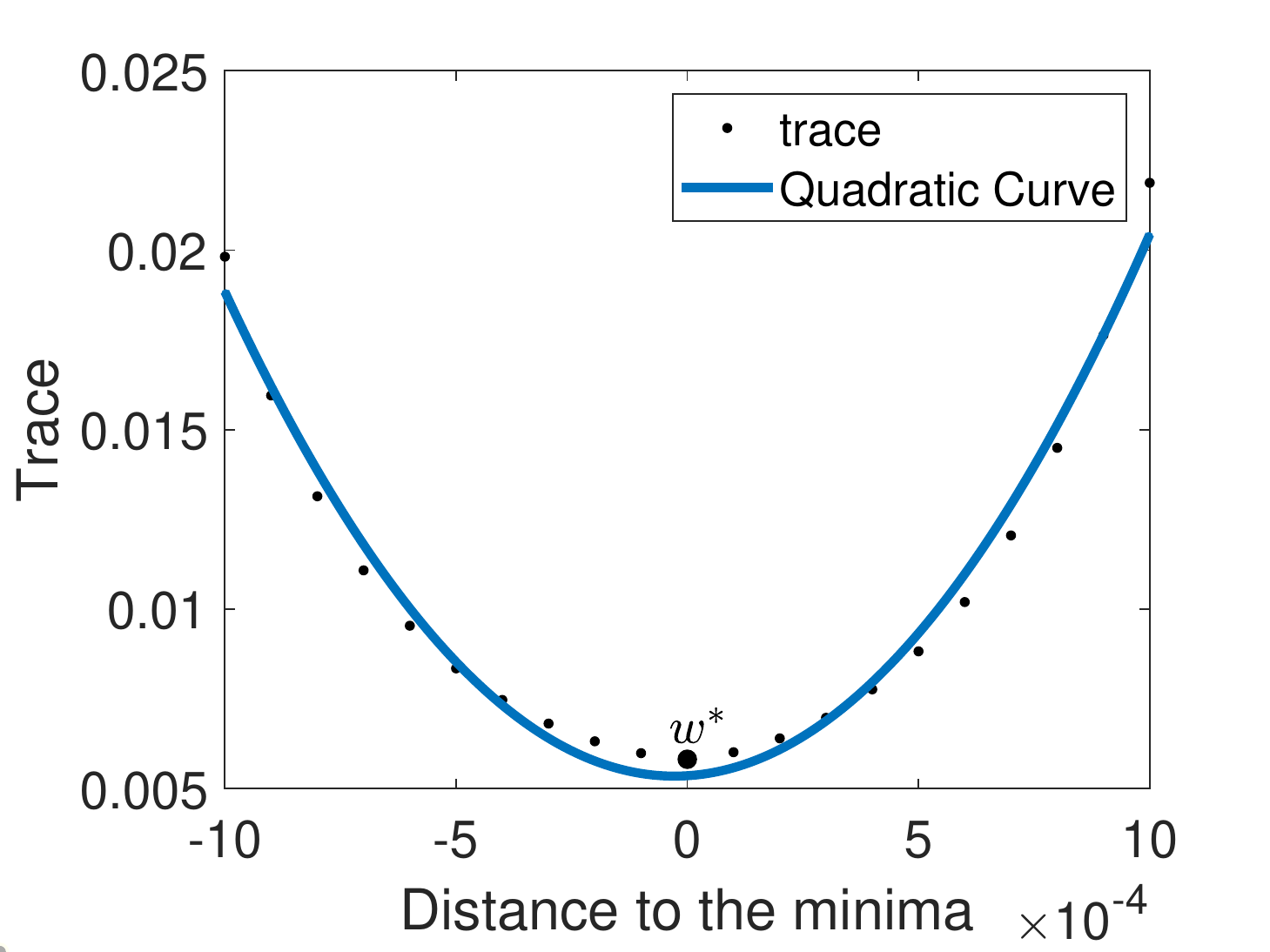}}
	\caption{Trace of covariance matrix of gradient noise in a region around local minimum $w^*$. $w^*$ is selected by running gradient descent with small learning rate till it converges. The number at horizontal axis shows the distance of the point away from $w^*$. \textbf{(a),(b):} Results for plain CNN. \textbf{(c),(d)}:Results for ResNet18.}
	\label{fig:2dmodel}
\end{figure}                        

We empirically observe the covariance matrix around the local minimum of training loss on deep neural networks. The results are shown in Figure.\ref{fig:2dmodel}. Readers can refer more details in Appendix 7.1. We have the following observations: (1) The traces of covariance matrices for the deep neural networks can be well approximated by quadratic curves, which supports Proposition \ref{propo1}.  (2) The minimum of the quadratic curve is nearly located at the local minimum $w^*$.  It indicates that the coefficient of the first-order term $\rho_{g,H}\approx 0$.  

Based on the fact that $\rho_{g,H}$ is not the determinant factor of the tail of the distribution in Eq.\ref{eq:densfull} and the observations in Figure.\ref{fig:2dmodel}, we consider a simplified form of $C(w)$ that $C(w)=\sigma_g+\sigma_H(w-w^*)^2$. 


\begin{corollary}\label{thm1}
	If $C(w)=\sigma_g+\sigma_H(w-w^*)^2$, the stationary distribution of power-law dynamic is
	\begin{equation}\label{eq:stdens}
	p(w)=\frac{1}{Z}(1+\sigma_H\sigma_g^{-1}(w-w^*)^2)^{-\kappa},
	\end{equation}
	where {\small$Z=\int_w(1+\sigma_H\sigma_g^{-1}(w-w^*)^2)^{-\kappa}dw$} is the normalization constant and $\kappa=\frac{H}{\eta\sigma_H}$ is the tail-index.
\end{corollary}
The distribution density in Eq.\ref{eq:stdens} is known as the power-law $\kappa$ distribution \cite{zhou2014kramers} (It is also named as $q$-Gaussian distribution in \cite{tsallis1996anomalous}). 
As $\kappa\rightarrow \infty$, the distribution density tends to be Gaussian, i.e.,  {\small$p(w)\propto\exp(-\frac{H(w-w^*)^2}{\eta\sigma_g})$}.  
Power-law $\kappa$ distribution becomes more heavy-tailed as $\kappa$ becomes smaller. Meanwhile, it produces higher probability to appear values far away from the center $w^*$. Intuitively, smaller $\kappa$ helps the dynamic to escape from local minima faster. 

In the approximation of dynamic of SGD, $\kappa$ equals the signal (i.e., $H(w^*)$) to noise (i.e., $\eta\sigma_H$) ratio of second-order derivative at $w^*$ in SGD, and $\kappa$ is linked with three factors:
(1) the curvature $H(w^*)$; (2) the fluctuation of the curvature over training data; (3) the hyper-parameters including $\eta$ and minibatch size $b$. Please note that $\sigma_H$ linearly decreases as the batch size $b$ increases. 


\subsection{Multivariate Power-Law Dynamic}
In this section, we extend the power-law dynamic to $d$-dimensional case.  
We first illustrate the covariance matrix $C(w)$ of gradient noise in SGD. We use the subscripts to denote the element in a vector or a matrix. It can be shown that {\small$C(w)=\Sigma_g(1+(w-w^*)^T\Sigma_H\Sigma_g^{-1}(w-w^*))$} where $\Sigma_g$ is the covariance matrix of $\tilde{g}(w^*)$ and {\small$\Sigma_H=Cov(\tilde{H}_{i}(w^*),\tilde{H}_j(w^*))$} is the covariance matrix of every two columns in {\small$\tilde{H}(w^*)$}. 
Here, we assume $\Sigma_g$ is isotropic (i.e., $\Sigma_g=\sigma_g\cdot I$) and {\small$Cov(\tilde{H}_{i}(w^*),\tilde{H}_j(w^*))$} are equal for all $i,j$. (Similarly as 1-dimensional case, we omit the first-order term $(w-w^*)$ in $C(w)$). Readers can refer Proposition 10 in Appendix 7.2 for the detailed derivation.

We suppose that the signal to noise ratio of $\tilde{H}(w^*)$ can be characterized by a scalar $\kappa$, i.e., $\eta\Sigma_H=\frac{1}{\kappa}\cdot H$. Then $C(w)$ can be written as {\small\begin{align}\label{eq:c}
	C(w)=\Sigma_g(1+\frac{1}{\eta\kappa}(w-w^*)^TH\Sigma_g^{-1}(w-w^*)).
	\end{align}}
\vspace{-0.4cm}
\begin{theorem}\label{prop2}
	If $w\in\mathbb{R}^d$ and  $C(w)$ has the form in Eq.(\ref{eq:c}). The stationary distribution density of power-law dynamic is 
	{\small	\begin{align}\label{eq:dens}
		p(w)=\frac{1}{Z}[1+\frac{1}{\eta\kappa}(w-w^*)^TH\Sigma_g^{-1}(w-w^*)]^{-{\kappa}}  
		\end{align}}where {\small$Z$} is the normalization constant and $\kappa$ satisfies $\eta\Sigma_H=\frac{1}{\kappa}\cdot H$.
\end{theorem}
\textbf{Remark:} The multivariate power-law $\kappa$ distribution (Eq.\ref{eq:dens}) is a natural extension of the 1-dimensional case. Actually, the assumptions on $\Sigma_g$ and  $\kappa$ can be replaced by just assuming $\Sigma_g, H, \Sigma_H$ are codiagonalized. Readers can refer Proposition 11 in Appendix 7.2 for the derivation.

\section{Escaping Efficiency of Power-law Dynamic}
\begin{wrapfigure}{r}{3.5cm}
	\vspace{-0.8cm}
	\centering
	\includegraphics[width=0.22\textwidth]{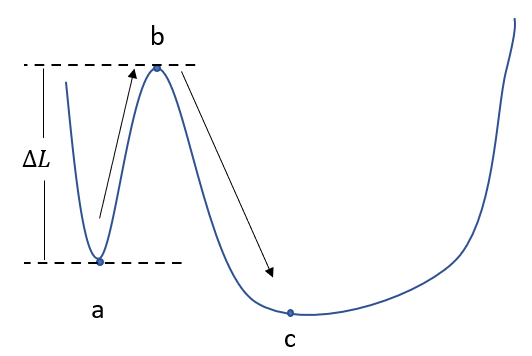}
	\vspace{-0.2cm}
	\caption{}
	\label{fig1}
	\vspace{-0.4cm}
\end{wrapfigure}
In this section, we analyze the escaping efficiency of power-law dynamic from local minima and its relation to generalization.
Specifically, we analyze the mean escaping time for $w_t$ to escape from a basin. As shown in Figure.\ref{fig1}, we suppose that there are two basins whose bottoms are denoted as $a$ and $c$ respectively and the saddle point $b$ is the barrier between two basins. The barrier height is denoted as $\Delta L=L(b)-L(a)$.

\begin{definition}Suppose $w_t$ starts at the local minimum $a$, we denote the time for $w_t$ to first reach the saddle point $b$ as $\inf\{t>0|w_0=a, w_t=b\}$. The mean escaping time $\tau$ is defined as $\tau=\mathbb{E}_{w_t}[\inf\{t>0|w_0=a, w_t=b\}]$.
\end{definition}
We first give the mean escaping time for 1-dimensional case in Lemma \ref{lemma} and then we give the mean escaping time for high-dimensional power-law dynamic in Theorem \ref{thm7}. To analyze the mean escaping time, we take the common \emph{low temperature} assumption \cite{xie2020diffusion,zhou2014kramers}, i.e., $\eta\sigma_g\ll\Delta L$. The assumption can be satisfied when the learning rate is small.

We suppose the basin $a$ is quadratic and the variance of noise has the form that  {\small$C(w)=\sigma_{g_a}+\sigma_{H_a}(w-a)^2$}, which can also be written as {\small$C(w)=\sigma_{g_a}+\frac{2\sigma_{H_a}}{H_a}(L(w)-L(a))$}.  
It means that the variance of gradient noise becomes larger as the loss becomes larger. 
The following lemma gives the mean escaping time of power-law dynamic for 1-dimensional case. 
\begin{lemma}\label{lemma}
	We suppose {\small$C(w)=\sigma_{g_a}+\frac{2\sigma_{H_a}}{H_a}(L(w)-L(a))$} on the whole escaping path from $a$ to $b$.	The mean escaping time of the 1-dimensional power-law dynamic is,
	{\small\begin{equation}
		\tau=\frac{2\pi}{(1-\frac{1}{2\kappa})\sqrt{{H_a}|H_b|}}\left(1+\frac{2}{\kappa\eta\sigma_{g_a}} \Delta L\right)^{\kappa-\frac{1}{2}},
		\end{equation}}where $\kappa=\frac{H_a}{\eta\sigma_{H_a}}>\frac{1}{2}$, $H_a$ and $H_b$ are the second-order derivatives of training loss at local minimum $a$ and at saddle point $b$, respectively.
\end{lemma}
The proof of Lemma \ref{lemma} is based on the results in \cite{zhou2014kramers}. We provide a full proof in Appendix 7.3. We summarize the mean escaping time of power-law dynamic and dynamics in previous works in Table \ref{tab:compa1}. Based on the results, we have the following discussions.
\begin{table*}[t]\footnotesize
	\centering
	\caption{Summary of related works and ours.  
		Here, we only show 1-dimensional result for escaping time in the table for all the three dynamics for ease of the presentation.
	}
	\label{tab:compa1}
	\begin{tabular}{cccc}
		\toprule[1.5pt]
		\textbf{Noise distribution}&\textbf{Dynamic}&\textbf{Stationary solution}&\textbf{Escaping time} \\
		\midrule[1.5pt]
		$\mathcal{N}(0,\sigma)$&Langevin&Gaussian&$\mathcal{O}\left(\frac{1}{\sqrt{H_a|H_b|}}\exp{\left(\frac{2\Delta L}{\eta\sigma}\right)}\right)$\\
		$\mathcal{S}(\alpha,\sigma)$&$\alpha$-stable&Heavy-tailed & $\mathcal{O}\left(\eta\alpha\cdot \left(\frac{|b-a|}{\eta\sigma}\right)^{\alpha}\right)$ \\
		$\mathcal{N}(0,\sigma_g+\sigma_H(w-w^*)^2)$&Power-law (ours)&Heavy-tailed&$\mathcal{O}\left(\frac{1}{\sqrt{H_a|H_b|}}(1+\frac{2}{\kappa}\frac{\Delta L}{\eta\sigma_g})^{\kappa-\frac{1}{2}}\right)$\\
		\bottomrule[1.5pt]
	\end{tabular}
\end{table*}

\textbf{Comparison with other dynamics:} (1) Both power-law dynamic and Langevin dynamic can escape sharp minima faster than flat minima, where the sharpness is measured by $H_a$ and larger $H_a$ corresponds to sharper minimum. Power-law dynamic improves the order of barrier height (i.e., $\Delta L$) from exponential to polynomial compared with Langevin dynamic, which implies a faster escaping efficiency of SGD to escape from deep basin. 
(2) The mean escaping time for $\alpha$-stable process is independent with the barrier height, but it is in polynomial order of the width of the basin (i.e., width=$|b-a|$). 
Compared with $\alpha$-stable process, the result for power-law dynamic is superior in the sense that it is also in polynomial order of the width (if $\Delta L\approx O(|b-a|^{2})$) and power-law dynamic does not rely on the infinite variance assumption.

Based on Lemma \ref{lemma}, we analyze the mean escaping time for $d$-dimensional case. Under the low temperature condition, the probability density concentrates only along the most possible escaping paths in the high-dimensional landscape. For rigorous definition of most possible escaping paths, readers can refer section 3 in \cite{xie2020diffusion}. For simplicity, we consider the case that there is only one most possible escaping path between basin a and basin c. Specifically, the Hessian at saddle point $b$ has only one negative eigenvalue and the most possible escaping direction is the direction corresponding to the negative eigenvalue of the Hessian at $b$.	

\begin{theorem}\label{thm7}
	Suppose $w\in\mathbb{R}^d$ and there is only one most possible path path between basin $a$ and basin $c$. The mean escaping time for power-law dynamic escaping from basin $a$ to basin $c$ is
	{\small\begin{equation}
		\tau=\frac{2\pi\sqrt{-\det(H_b)}}{(1-\frac{d}{2\kappa})\sqrt{\det(H_a)}}\frac{1}{|H_{be}|}\left(1+\frac{1}{\eta\kappa\sigma_{e}} \Delta L\right)^{\kappa-\frac{1}{2}},
		\end{equation}}where $e$ indicates the most possible escaping direction, $H_{be}$ is the only negative eigenvalue of $H_b$, $\sigma_{e}$ is the eigenvalue of $\Sigma_{g_a}$ that corresponds to the escaping direction, $\Delta L=L(b)-L(a)$, and $\det(\cdot)$ is the determinant of a matrix.
\end{theorem}
\textbf{Remark:} In $d$-dimensional case, the flatness is measured by $\det(H_a)$. If $H_a$ has zero eigenvalues, we can replace $H_a$ by  $H_a^{+}$ in above theorem, where $H_a^{+}$ is obtained by projecting $H_a$ onto the subspace composed by the eigenvectors corresponding to the positive eigenvalues of $H_a$. 
This is because by Taylor expansion, the loss $L(w)$ only depends on the positive eigenvalues and the corresponding eigenvectors of $H_a$, i.e., {\small$L(w)=L(a)+\frac{1}{2}(w-a)^TH_a(w-a)=L(a)+\frac{1}{2}(\mathbb{P}(w-a))^T\Lambda_{H_a^+}\mathbb{P}(w-a)$}, where {\small$\Lambda_{H_a^+}$} is a diagonal matrix composed by non-zero eigenvalues of $H_a$ and the operator $\mathbb{P}(\cdot)$ operates the vector to the subspace corresponding to non-zero eigenvalues of $H_a$. 
The next theorem give an upper bound of the generalization error of the stationary distribution of power-law dynamic, which shows that flatter minimum has smaller generalization error.
\begin{theorem}\label{thm9}
	Suppose that $w\in\mathbb{R}^d$ and $\kappa>\frac{d}{2}$. 
	For $\delta>0$, with probability at least $1-\delta$, the stationary distribution of power-law dynamic has the following generalization error bound,
	{\begin{align*}
		\mathbb{E}_{w\sim p(w), x\sim \mathcal{P}(x)}\ell(w,x)\leq\mathbb{E}_{w\sim p(w)}L(w)+\sqrt{\frac{KL(p||p')+\log\frac{1}{\delta}+\log{n}+2}{n-1}},
		\end{align*}}where {$KL(p||p')\leq\frac{1}{2}\log\frac{\det(H)}{\det(\Sigma_g)}+\frac{Tr(\eta\Sigma_gH^{-1})-2d}{4\left(1-\frac{1}{\kappa}\left(\frac{d}{2}-1\right)\right)}+\frac{d}{2}\log\frac{2}{\eta}$}, $p(w)$ is the stationary distribution of $d$-dimensional power-law dynamic, $p'(w)$ is a prior distribution which is selected to be standard Gaussian distribution,   and $\mathcal{P}(x)$ is the underlying distribution of data $x$, $\det(\cdot)$ and $Tr(\cdot)$ are the determinant and trace of a matrix, respectively.
\end{theorem}
Theorem \ref{thm9} shows that if {\small$2d>Tr(\eta\Sigma_gH^{-1})$}, the generalization error is decreasing as $\det(H)$ decreases, which indicates that flatter minimum has smaller generalization error. Moreover, if {\small$2d>Tr(\eta\Sigma_gH^{-1})$}, the generalization error is decreasing as $\kappa$ increases. When $\kappa\rightarrow\infty$, the generalization error tends to that for Langevin dynamic. Combining the mean escaping time and the generalization error bound, we can conclude that state-dependent noise makes SGD escape from sharp minima faster and implicitly tend to learn a flatter model which generalizes better.

\section{Experiments}
In this section, we conduct experiments to verify the theoretical results. We first study the fitness between parameter distribution trained by SGD and power-law $\kappa$ distribution. Then we compare the escaping behavior for power-law dynamic, Langevin dynamic and SGD.
\subsection{Fitting Parameter Distribution using Power-Law Distribution}
We investigate the distribution of parameters trained by SGD on deep neural networks and use power-law $\kappa$ distribution to fit the parameter distribution. We first use SGD to train various types of deep neural networks till it converge. For each network, we run SGD with different minibatch sizes over the range $\{64,256,1024\}$. For the settings of other hyper-parameters, readers can refer Appendix 7.5.2.
We plot the distribution of model parameters at the same layer using histogram. 
Next, we use power-law $\kappa$ distribution to fit the distribution of the parameters and estimate the value of $\kappa$ via the embedded function "$TsallisQGaussianDistribution[ ]$"
in Mathematica software. 

We show results for LeNet-5 with MNIST dataset and ResNet-18 with CIFAR10 dataset \cite{lecun2015lenet,he2016deep} in this section,  and put results for other network architectures in Appendix 7.5.2. 
We have the following observations: (1) The distribution of the parameter trained by SGD can be well fitted by power-law $\kappa$ distribution (blue curve). (2) As the minibatch size becomes larger, $\kappa$ becomes larger. It is because the noise $\sigma_H$ linearly decreases as minibatch size becomes larger and $\kappa=\frac{H}{\eta\sigma_H}$. (3) For results on LeNet-5, as $\kappa$ becomes larger, the test accuracy becomes lower. Meanwhile, all the training losses are lower than $10^{-3}$ on LeNet-5. It indicates that $\kappa$ also plays a role as indicator of generalization. These results are consistent with the theory in Section 4.

\subsection{Comparison on Escaping Efficiency}\label{sec5.2}
We use a 2-dimensional model to simulate the escaping efficiency from minima for power-law dynamic, Langevin dynamic and SGD. We design a non-convex 2-dimensional function written as {$L(w) =\frac{1}{n} \sum_{i=1}^n \ell(w-x_i)$}, where {$\ell(w) = 15 \sum_{j=1}^2 |w_j-1|^{2.5}\cdot|w_j+1|^{3}$} and training data $x_i \sim \mathcal{N}(0,0.01I_2)$. 
We regard the following optimization iterates as the numerical discretization of the power-law dynamic, {$	w_{t+1} =w_{t}-\eta g(w_{t}) + \eta\lambda_2 \sqrt{1+\lambda_1 L(w_t)} \odot \xi $,} 
where $\xi \sim \mathcal{N}(0,I_2)$, $\lambda_1, \lambda_2$ are two hyper-parameters and $\odot$ stands for Hadamard product. Note that if we set $\lambda_1=0$, it can be regarded as discretization of Langevin dynamic.
We set learning rate $\eta = 0.025$, and we take $500$ iterations in each training. In order to match the trace of covariance matrix of stochastic gradient at minimum point $w^*$ with the methods above, $\lambda_2$ is chosen to satisfy {$Tr(Cov(\lambda_2\xi)) = Tr(Cov(g(w^*)))$}. %

\begin{figure}[t]
	\centering
	\subfigure[LeNet-5 conv layer 2]{\label{fig:pl1}\includegraphics[width=0.49\textwidth]{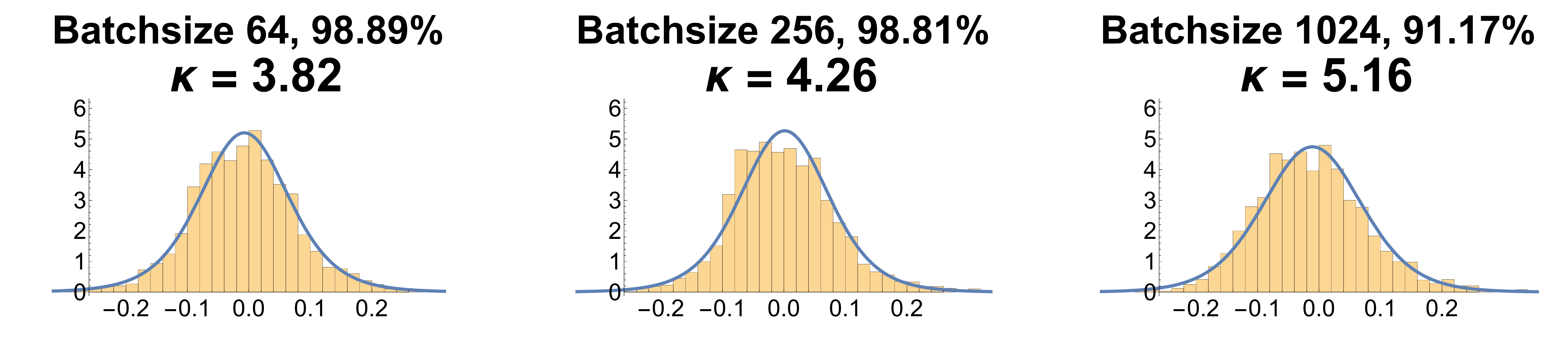}}
	\subfigure[ResNet-18 layer 1]{\label{fig:pl2}\includegraphics[width=0.49\textwidth]{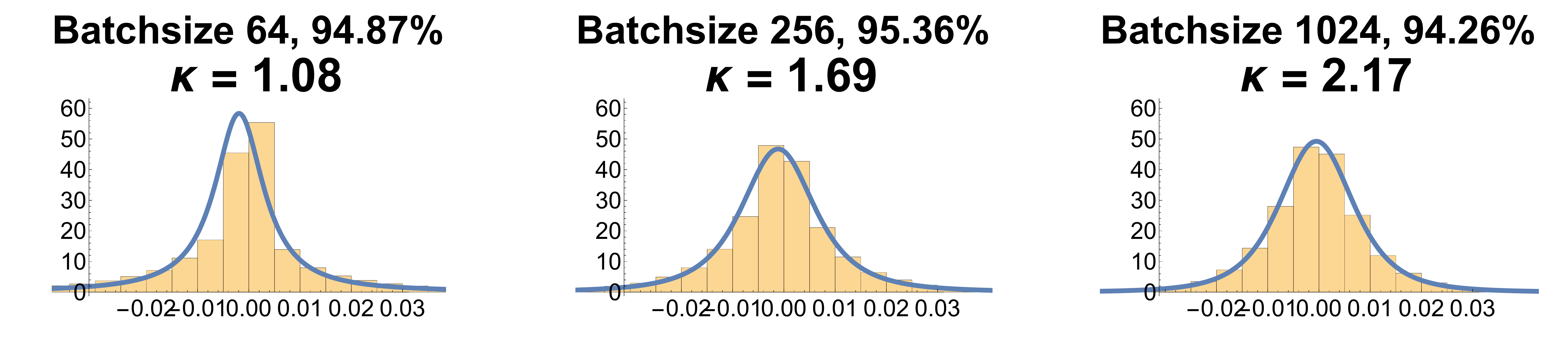}}
	\caption{Approximating distribution of parameters (trained by SGD) by power-law dynamic. The percentage on the right side of batchsize denotes final test accuracy respectively. \textbf{(a):} Results for LeNet-5. \textbf{(b)}:Results for ResNet-18. }
	\label{fig:powerdis}
\end{figure}

\begin{figure}[h]
	\centering
	\subfigure[2-D loss]{\label{fig:loss}\includegraphics[width=0.25\textwidth]{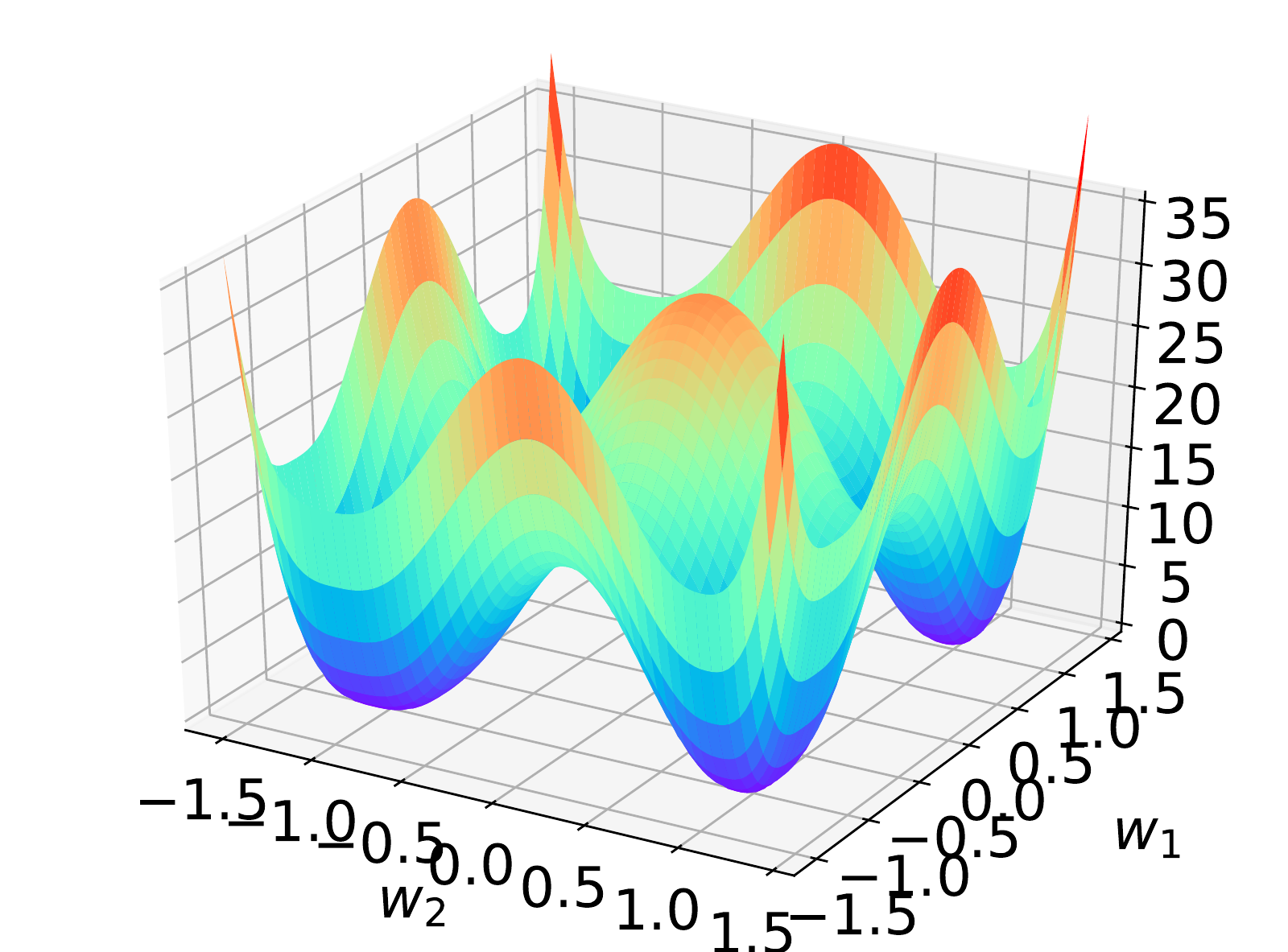}}
	\subfigure[Trace for covariance]{\label{fig:cov1}\includegraphics[width=0.25\textwidth]{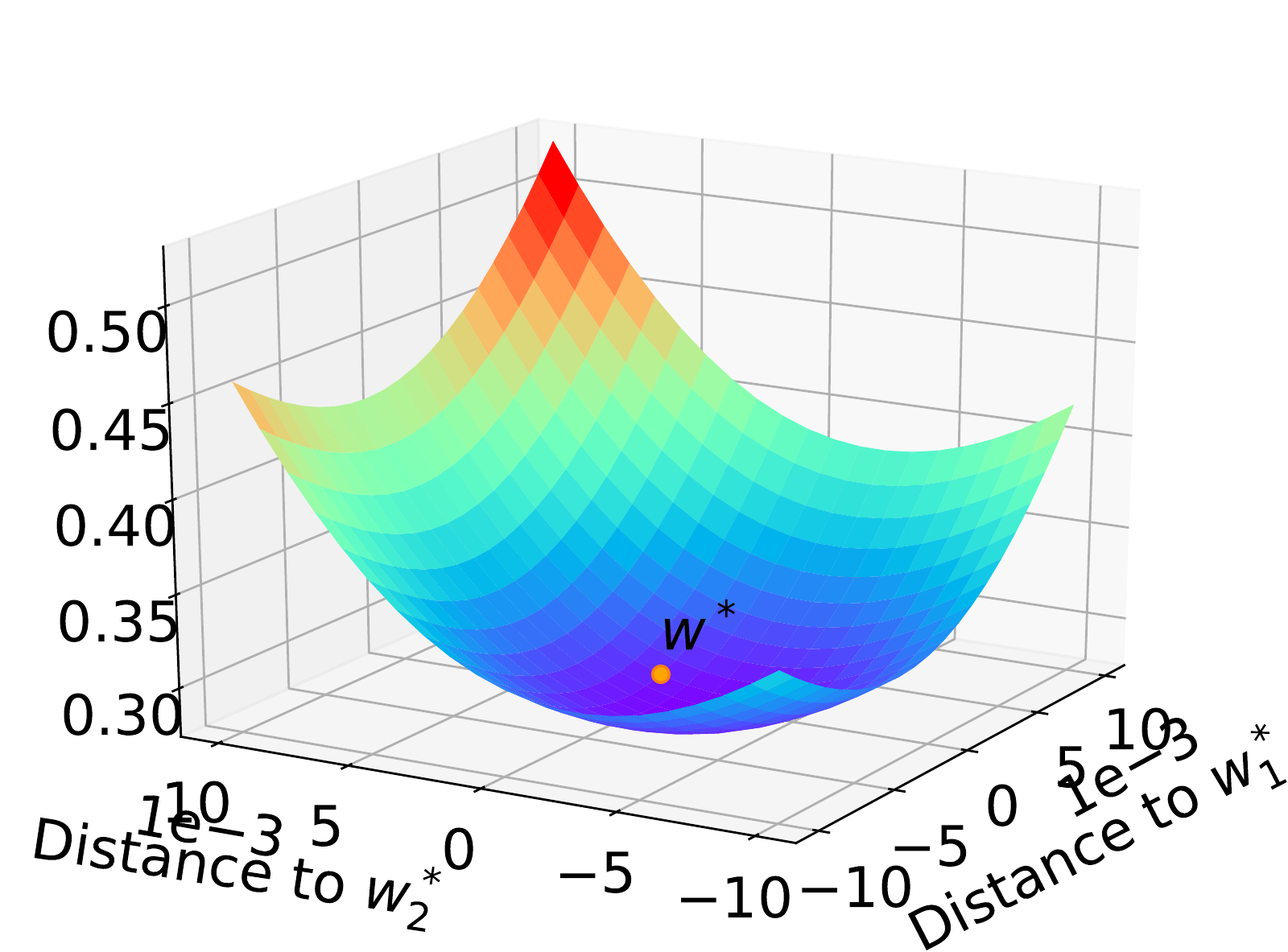}}
	\subfigure[Success Rate (sharp)]{\label{fig:succn1}\includegraphics[width=0.23\textwidth]{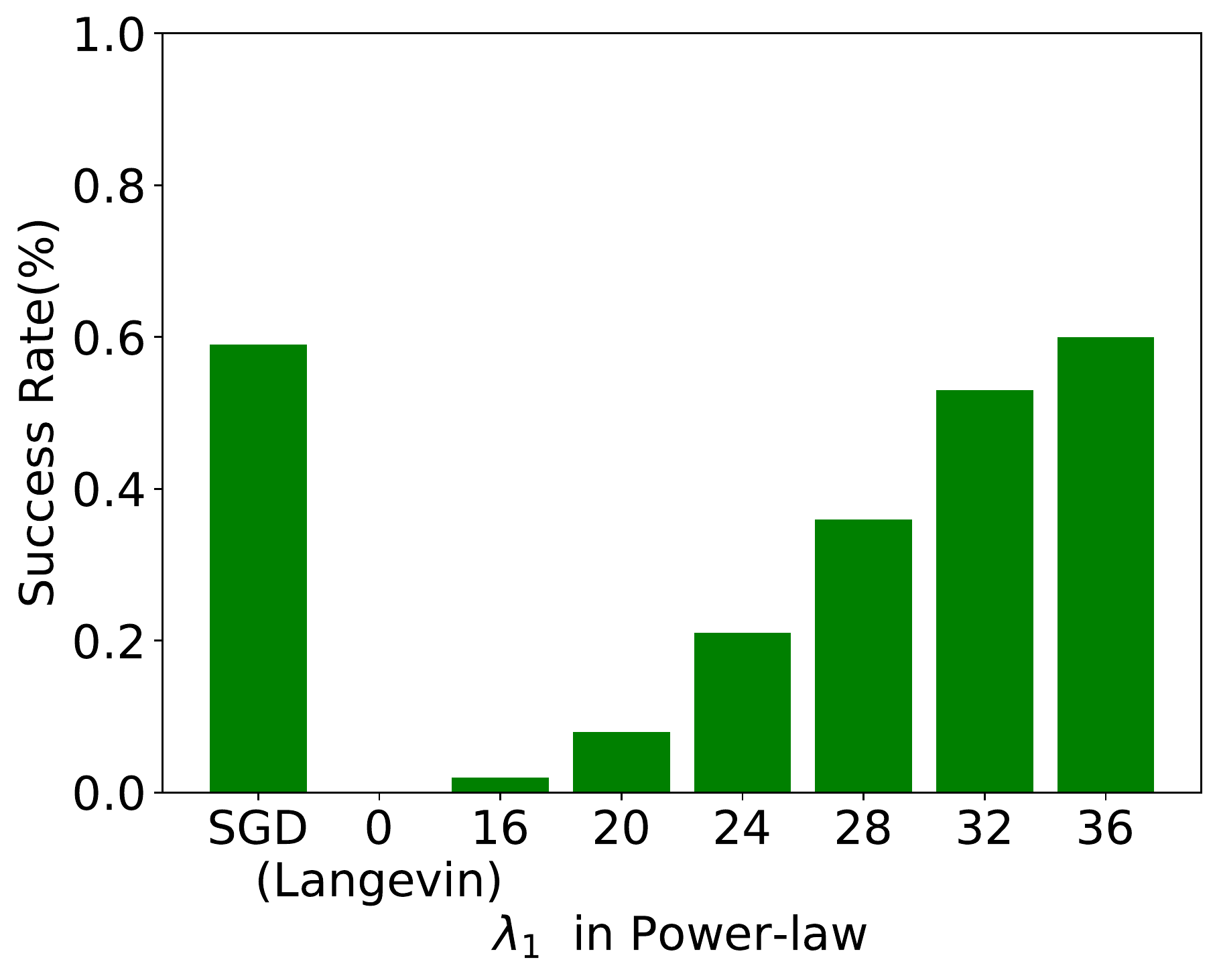}}
	\subfigure[Success Rate (flat)]{\label{fig:succ21}\includegraphics[width=0.23\textwidth]{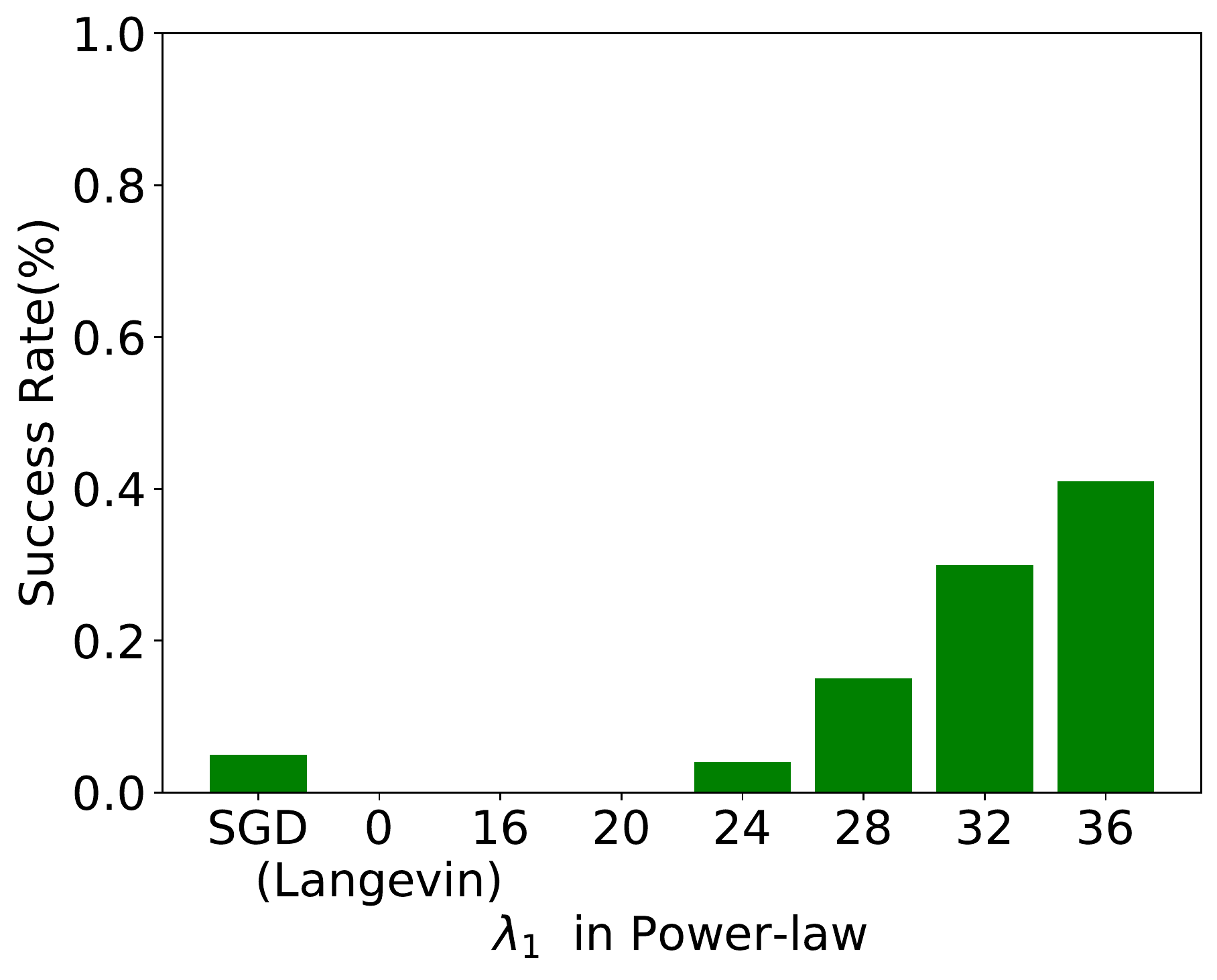}}
	\caption{ \textbf{(a):}Loss surface of $L(w)$ for 2-D model. \textbf{(b)}:Trace of covariance matrix around minimum $(1,1)$. \textbf{(c)/(d):} Success rate of escaping from the basin of $L(w)$ / $0.9L(w)$ in repeated 100 runs.}
	\label{fig:2dmodel2}
\end{figure}

We compare the success rate of escaping for power-law dynamic, Langevin dynamic and SGD by repeating the experiments 100 times. To analyze the noise term $\lambda_1$, we choose different $\lambda_1$ and evaluate corresponding success rate of escaping, as shown in Figure.\ref{fig:succn1}. The results show that: (1) there is a positive correlation between $\lambda_1$ and the success rate of escaping; (2) power-law dynamic can mimic the escaping efficiency of SGD, while Langevin dynamic can not. 	We then scale the loss function by $0.9$ to make the minima flatter and repeat all the algorithms under the same setting. The success rate for the scaled loss function is shown in Figure.\ref{fig:succ21}. We can observe that all dynamics escape flatter minima slower. 
\section{Conclusion}
In this work, we study the dynamic of SGD via investigating state-dependent variance of the stochastic gradient. We propose power-law dynamic with state-dependent diffusion to approximate the dynamic of SGD. We analyze the escaping efficiency from local minima and the PAC-Bayes generalization error bound for power-law dynamic.  Results indicate that state-dependent noise helps SGD escape from poor local minima faster and generalize better. We present direct empirical evidence to support our theoretical findings.
This work may motivate many interesting research topics, for example, non-Gaussian state-dependent noise, new types of state-dependent regularization tricks in deep learning algorithms and more accurate characterization about the loss surface of deep neural networks. We will investigate these topics in future work.

\bibliographystyle{plain}
\bibliography{pbn}

\begin{thebibliography}{10}

\bibitem{bottou2008tradeoffs}
L{\'e}on Bottou and Olivier Bousquet.
\newblock The tradeoffs of large scale learning.
\newblock In {\em Advances in neural information processing systems}, pages
  161--168, 2008.

\bibitem{chaudhari2018stochastic}
Pratik Chaudhari and Stefano Soatto.
\newblock Stochastic gradient descent performs variational inference, converges
  to limit cycles for deep networks.
\newblock In {\em 2018 Information Theory and Applications Workshop (ITA)},
  pages 1--10. IEEE, 2018.

\bibitem{choromanska2015loss}
Anna Choromanska, Mikael Henaff, Michael Mathieu, G{\'e}rard~Ben Arous, and
  Yann LeCun.
\newblock The loss surfaces of multilayer networks.
\newblock In {\em Artificial intelligence and statistics}, pages 192--204,
  2015.

\bibitem{draxler2018essentially}
Felix Draxler, Kambis Veschgini, Manfred Salmhofer, and Fred Hamprecht.
\newblock Essentially no barriers in neural network energy landscape.
\newblock In {\em International Conference on Machine Learning}, pages
  1309--1318, 2018.

\bibitem{guo2014power}
Ran Guo and Jiulin Du.
\newblock Are power-law distributions an equilibrium distribution or a
  stationary nonequilibrium distribution?
\newblock {\em Physica A: Statistical Mechanics and its Applications},
  406:281--286, 2014.

\bibitem{gurbuzbalaban2020heavy}
Mert Gurbuzbalaban, Umut Simsekli, and Lingjiong Zhu.
\newblock The heavy-tail phenomenon in sgd.
\newblock {\em arXiv preprint arXiv:2006.04740}, 2020.

\bibitem{haochen2020shape}
Jeff~Z HaoChen, Colin Wei, Jason~D Lee, and Tengyu Ma.
\newblock Shape matters: Understanding the implicit bias of the noise
  covariance.
\newblock {\em arXiv preprint arXiv:2006.08680}, 2020.

\bibitem{he2016dual}
Di~He, Yingce Xia, Tao Qin, Liwei Wang, Nenghai Yu, Tie-Yan Liu, and Wei-Ying
  Ma.
\newblock Dual learning for machine translation.
\newblock In {\em Advances in neural information processing systems}, pages
  820--828, 2016.

\bibitem{he2019control}
Fengxiang He, Tongliang Liu, and Dacheng Tao.
\newblock Control batch size and learning rate to generalize well: Theoretical
  and empirical evidence.
\newblock In {\em Advances in Neural Information Processing Systems}, pages
  1141--1150, 2019.

\bibitem{he2019asymmetric}
Haowei He, Gao Huang, and Yang Yuan.
\newblock Asymmetric valleys: Beyond sharp and flat local minima.
\newblock In {\em Advances in Neural Information Processing Systems}, pages
  2549--2560, 2019.

\bibitem{he2015delving}
Kaiming He, Xiangyu Zhang, Shaoqing Ren, and Jian Sun.
\newblock Delving deep into rectifiers: Surpassing human-level performance on
  imagenet classification.
\newblock In {\em Proceedings of the IEEE international conference on computer
  vision}, pages 1026--1034, 2015.

\bibitem{he2016deep}
Kaiming He, Xiangyu Zhang, Shaoqing Ren, and Jian Sun.
\newblock Deep residual learning for image recognition.
\newblock In {\em Proceedings of the IEEE conference on computer vision and
  pattern recognition}, pages 770--778, 2016.

\bibitem{he2018differential}
Li~He, Qi~Meng, Wei Chen, Zhi-Ming Ma, and Tie-Yan Liu.
\newblock Differential equations for modeling asynchronous algorithms.
\newblock In {\em Proceedings of the 27th International Joint Conference on
  Artificial Intelligence}, pages 2220--2226, 2018.

\bibitem{hodgkinson2020multiplicative}
Liam Hodgkinson and Michael~W Mahoney.
\newblock Multiplicative noise and heavy tails in stochastic optimization.
\newblock {\em arXiv preprint arXiv:2006.06293}, 2020.

\bibitem{hu2019diffusion}
Wenqing Hu, Chris~Junchi Li, Lei Li, and Jian-Guo Liu.
\newblock On the diffusion approximation of nonconvex stochastic gradient
  descent.
\newblock {\em Annals of Mathematical Sciences and Applications}, 4(1):3--32,
  2019.

\bibitem{keskar2016large}
Nitish~Shirish Keskar, Dheevatsa Mudigere, Jorge Nocedal, Mikhail Smelyanskiy,
  and Ping Tak~Peter Tang.
\newblock On large-batch training for deep learning: Generalization gap and
  sharp minima.
\newblock {\em arXiv preprint arXiv:1609.04836}, 2016.

\bibitem{lecun2015lenet}
Yann LeCun et~al.
\newblock Lenet-5, convolutional neural networks.
\newblock {\em URL: http://yann. lecun. com/exdb/lenet}, 20(5):14, 2015.

\bibitem{li2018over}
Dawei Li, Tian Ding, and Ruoyu Sun.
\newblock Over-parameterized deep neural networks have no strict local minima
  for any continuous activations.
\newblock {\em arXiv preprint arXiv:1812.11039}, 2018.

\bibitem{li2018visualizing}
Hao Li, Zheng Xu, Gavin Taylor, Christoph Studer, and Tom Goldstein.
\newblock Visualizing the loss landscape of neural nets.
\newblock In {\em Advances in Neural Information Processing Systems}, pages
  6389--6399, 2018.

\bibitem{li2017stochastic}
Qianxiao Li, Cheng Tai, et~al.
\newblock Stochastic modified equations and adaptive stochastic gradient
  algorithms.
\newblock In {\em Proceedings of the 34th International Conference on Machine
  Learning-Volume 70}, pages 2101--2110. JMLR. org, 2017.

\bibitem{liu2018toward}
Tianyi Liu, Zhehui Chen, Enlu Zhou, and Tuo Zhao.
\newblock Toward deeper understanding of nonconvex stochastic optimization with
  momentum using diffusion approximations.
\newblock {\em arXiv preprint arXiv:1802.05155}, 2018.

\bibitem{mahoney2019traditional}
Michael Mahoney and Charles Martin.
\newblock Traditional and heavy tailed self regularization in neural network
  models.
\newblock In {\em International Conference on Machine Learning}, pages
  4284--4293, 2019.

\bibitem{mandt2017stochastic}
Stephan Mandt, Matthew~D Hoffman, and David~M Blei.
\newblock Stochastic gradient descent as approximate bayesian inference.
\newblock {\em The Journal of Machine Learning Research}, 18(1):4873--4907,
  2017.

\bibitem{mcallester1999pac}
David~A McAllester.
\newblock Pac-bayesian model averaging.
\newblock In {\em Proceedings of the twelfth annual conference on Computational
  learning theory}, pages 164--170, 1999.

\bibitem{rakhlin2012making}
Alexander Rakhlin, Ohad Shamir, and Karthik Sridharan.
\newblock Making gradient descent optimal for strongly convex stochastic
  optimization.
\newblock In {\em Proceedings of the 29th International Coference on
  International Conference on Machine Learning}, pages 1571--1578, 2012.

\bibitem{csimcsekli2019heavy}
Umut {\c{S}}im{\c{s}}ekli, Mert G{\"u}rb{\"u}zbalaban, Thanh~Huy Nguyen,
  Ga{\"e}l Richard, and Levent Sagun.
\newblock On the heavy-tailed theory of stochastic gradient descent for deep
  neural networks.
\newblock {\em arXiv preprint arXiv:1912.00018}, 2019.

\bibitem{simsekli2019tail}
Umut Simsekli, Levent Sagun, and Mert Gurbuzbalaban.
\newblock A tail-index analysis of stochastic gradient noise in deep neural
  networks.
\newblock In {\em International Conference on Machine Learning}, pages
  5827--5837, 2019.

\bibitem{smith2017bayesian}
Samuel~L Smith and Quoc~V Le.
\newblock A bayesian perspective on generalization and stochastic gradient
  descent.
\newblock {\em arXiv preprint arXiv:1710.06451}, 2017.

\bibitem{tsallis1995anomalous}
Constantino Tsallis and Dirk~Jan Bukman.
\newblock Anomalous diffusion in the presence of external forces: exact
  time-dependent solutions and entropy.
\newblock {\em arXiv preprint cond-mat/9511007}, 1995.

\bibitem{tsallis1996anomalous}
Constantino Tsallis and Dirk~Jan Bukman.
\newblock Anomalous diffusion in the presence of external forces: Exact
  time-dependent solutions and their thermostatistical basis.
\newblock {\em Physical Review E}, 54(3):R2197, 1996.

\bibitem{van1992stochastic}
Nicolaas~Godfried Van~Kampen.
\newblock {\em Stochastic processes in physics and chemistry}, volume~1.
\newblock Elsevier, 1992.

\bibitem{vaswani2017attention}
Ashish Vaswani, Noam Shazeer, Niki Parmar, Jakob Uszkoreit, Llion Jones,
  Aidan~N Gomez, {\L}ukasz Kaiser, and Illia Polosukhin.
\newblock Attention is all you need.
\newblock In {\em Advances in neural information processing systems}, pages
  5998--6008, 2017.

\bibitem{wu2019noisy}
Jingfeng Wu, Wenqing Hu, Haoyi Xiong, Jun Huan, Vladimir Braverman, and
  Zhanxing Zhu.
\newblock On the noisy gradient descent that generalizes as sgd.
\newblock {\em arXiv preprint arXiv:1906.07405}, 2019.

\bibitem{wu2019multiplicative}
Jingfeng Wu, Wenqing Hu, Haoyi Xiong, Jun Huan, and Zhanxing Zhu.
\newblock The multiplicative noise in stochastic gradient descent:
  Data-dependent regularization, continuous and discrete approximation.
\newblock {\em CoRR}, 2019.

\bibitem{xiao2017fashion}
Han Xiao, Kashif Rasul, and Roland Vollgraf.
\newblock Fashion-mnist: a novel image dataset for benchmarking machine
  learning algorithms.
\newblock {\em arXiv preprint arXiv:1708.07747}, 2017.

\bibitem{xie2020diffusion}
Zeke Xie, Issei Sato, and Masashi Sugiyama.
\newblock A diffusion theory for deep learning dynamics: Stochastic gradient
  descent escapes from sharp minima exponentially fast.
\newblock {\em arXiv preprint arXiv:2002.03495}, 2020.

\bibitem{zhang2018energy}
Yao Zhang, Andrew~M Saxe, Madhu~S Advani, and Alpha~A Lee.
\newblock Energy--entropy competition and the effectiveness of stochastic
  gradient descent in machine learning.
\newblock {\em Molecular Physics}, 116(21-22):3214--3223, 2018.

\bibitem{zhou2019toward}
Mo~Zhou, Tianyi Liu, Yan Li, Dachao Lin, Enlu Zhou, and Tuo Zhao.
\newblock Toward understanding the importance of noise in training neural
  networks.
\newblock In {\em International Conference on Machine Learning}, 2019.

\bibitem{zhou2014kramers}
Yanjun Zhou and Jiulin Du.
\newblock Kramers escape rate in overdamped systems with the power-law
  distribution.
\newblock {\em Physica A: Statistical Mechanics and its Applications},
  402:299--305, 2014.

\bibitem{zhu2019anisotropic}
Zhanxing Zhu, Jingfeng Wu, Bing Yu, Lei Wu, and Jinwen Ma.
\newblock The anisotropic noise in stochastic gradient descent: Its behavior of
  escaping from sharp minima and regularization effects.
\newblock In {\em Proceedings of International Conference on Machine Learning},
  pages 7654--7663, 2019.

\end{thebibliography}

\section{Appendix}
\subsection{Power-law Dynamic and Stationary Distribution}
\begin{theorem} (Corollary 3 in main paper)
	If $C(w)=\sigma_g+\sigma_H(w-w^*)^2$, the stationary distribution density of power-law dynamic is
	\begin{equation}
	p(w)=\frac{1}{Z}(1+\sigma_H\sigma_g^{-1}(w-w^*)^2)^{-\kappa},
	\end{equation}
	where $Z=\int_w(1+\sigma_H\sigma_g^{-1}(w-w^*)^2)^{-\kappa}dw$ is the normalization constant and $\kappa=\frac{H}{\eta\sigma_H}$ is the tail-index.
\end{theorem}
\textit{Proof:} According to the Fokker-Planck equation, $p(w)$ satisfies
\begin{align*}
0&=\nabla p(w)g(w)+\frac{\eta}{2}\cdot\nabla\cdot\left(C(w)\nabla p(w)\right)\\
&=\nabla (p(w)\cdot \nabla L(w))+\frac{\eta}{2} \nabla\cdot (\sigma_g+\frac{2\sigma_H}{H}(L(w)-L(w^*)))\nabla p(w) \\
&=\nabla\cdot \frac{\eta}{2}C(w) (1+\frac{2\sigma_H}{H\sigma_g}(L(w)-L(w^*)))^{\frac{H}{-\eta\sigma_H}}\nabla (1+\frac{2\sigma_H}{H\sigma_g}(L(w)-L(w^*)))^{\frac{H}{\eta\sigma_H}}p(w)
\end{align*}
Because the left side equals zero, we have $(1+\frac{2\sigma_H}{H\sigma_g}(L(w)-L(w^*)))^{\frac{H}{\eta\sigma_H}}p(w)$ equals to constant. So $p(w)\propto (1+\frac{2\sigma_H}{H\sigma_g}(L(w)-L(w^*)))^{\frac{H}{-\eta\sigma_H}}$. So we can get the conclusion in the theorem. $\Box$

\begin{figure}[h]
	\centering
	\includegraphics[width=0.4\textwidth]{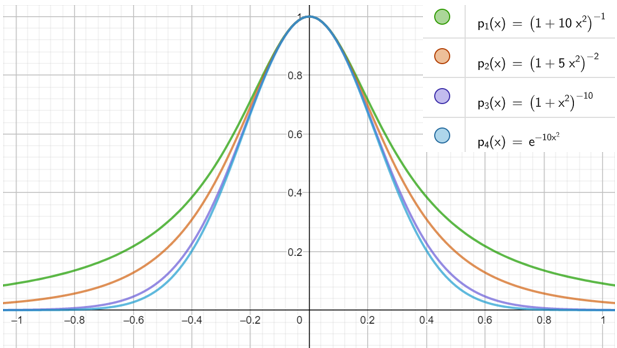}
	\caption{Probability density for power-law dynamic.}
	\label{fig111}
\end{figure}

We plot the un-normalized distribution density for 1-dimensional power-law dynamics with different $\kappa$ in Figure \ref{fig111}. For the four curves, we set $\beta=10$. We set $\kappa=1, 0.5, 0.1, 0$ and use green, red, purple and blue line to illustrate their corresponding density function, respectively. When $\kappa=0$, it is Gaussian distribution. From the figure, we can see that the tail for power-law $\kappa$-distribution is heavier than Gaussian distribution.

Actually, for any given time $t$, the distribution $p(w,t)$ for $w_t$ that satisfies power-law dynamic has analytic form, i.e.,  $p(w,t)\propto (1+\frac{H}{\eta\kappa\sigma(t)}(w-w(t))^2)^{-\kappa}$, where $w(t)=w^*+(w_0-w^*)e^{-Ht}$ and $\sigma(t)$ is a function of $\sigma_g$ and $t$. Readers can refer Eq.18 - Eq.23 in \cite{tsallis1995anomalous} for the detailed expression. 

As for the proof of Theorem 2 in the main paper, it is similar with the proof of corollary 3. Readers can verify that Eq.6 satisfies the Fokker-Planck equation and we omit its proof here. 

\subsection{SGD and Multivariate Power-law Dynamic}
The following proposition shows the covariance of stochastic gradient in SGD in $d$-dimensional case. We use the subscripts to denote the elements in a vector or a matrix.
\begin{proposition}\label{prop10}
	For $w\in\mathbb{R}^d$, we use $C(w)$ to denote the covariance matrix of stochastic gradient $\tilde{g}(w)=\tilde{g}(w^*)+\widetilde{H}(w-w^*)$ and $\Sigma$ to denote the covariance matrix of $\tilde{g}(w^*)$. If $Cov(\tilde{g}_i(w^*),\widetilde{H}_{jk})=0,\forall i,j,k$, we have \begin{align}
	C_{ij}(w)=\Sigma_{ij}+(w-w^*)^TA^{(ij)}(w-w^*), \label{cov}
	\end{align}where $\Sigma_{ij}=Cov(\tilde{g}_i(w^*),\tilde{g}_j(w^*))$, $A^{(ij)}$ is a $d\times d$ matrix with elements $A^{(ij)}_{ab}=Cov(\tilde{H}_{ia},\tilde{H}_{jb})$ with $a\in[d], b\in[d]$. 
\end{proposition}
Eq.\ref{cov} can be obtained by directly calculating the covariance of $\tilde{g}_i(w)$ and $\tilde{g}_j(w)$ where $\tilde{g}_i(w)=\tilde{g}_i(w^*)+\sum_{a=1}^d\tilde{H}_{ia}(w_a-w_a^*)$, $\tilde{g}_j(w)=\tilde{g}_j(w^*)+\sum_{b=1}^d\tilde{H}_{jb}(w_b-w_b^*)$. 

In order to get a analytic tractable form of $C(w)$, we make the following assumptions: (1) If $\Sigma_{ij}=0$, $A^{(ij)}$ is a zero matrix; (2) For $\Sigma_{ij}\neq 0$, $\frac{A^{(ij)}}{\Sigma_{ij}}$ are equal for all $i\in[d],j\in[d]$. The first assumption is reasonable because both $\Sigma_{ij}$ and $A^{(ij)}$ reflect the dependence of the derivatives along the $i$-th direction and $j$-th direction. Let $\Sigma_H=\frac{A^{(ij)}}{\Sigma_{ij}}$, $C(w)$ can be written as $C(w)=\Sigma_g(1+(w-w^*)^T\Sigma_H(w-w^*))$. The $d$-dimensional power-law dynamic is written as
\begin{equation}
dw_t=-H(w-w^*)dt+\sqrt{\eta C(w)} dB_t,   
\end{equation}where $C(w)=\Sigma_g(1+(w-w^*)^T\Sigma_H(w-w^*))$ which is a symmetric positive definite matrix that $C(w)^{1/2}$ exists. The following proposition shows the stationary distribution of the $d$-dimensional power-law dynamic.
\begin{proposition}
	Suppose $\Sigma_g, \Sigma_H, H$ are codiagonalizable, i.e., there exist orthogonal matrix $Q$ and diagonal matrices $\Lambda, \Gamma, \Pi$ to satisfy $\Sigma_g=Q^T\Lambda Q, {\Sigma_H}=Q^T\Gamma Q, H=Q^T\Pi Q$. Then, the stationary distribution of power-law dynamic is
	\begin{align}
	p(w)=\frac{1}{Z}(1+(w-w^*)^T\Sigma_H(w-w^*))^{-\kappa},
	\end{align}where $Z$ is the normalization constant and $\kappa=\frac{Tr(H)}{\eta Tr(\Sigma_H\Sigma_g)}$.
\end{proposition}
\textit{Proof:}
Under the codiagonalization assumption on $\Sigma_g, \Sigma_H, H$, Eq.15 can be rewritten as $dv_t=-\Pi v_tdt+\sqrt{\eta\Lambda(1+v_t^T\Gamma v_t)}dB_t$ if we let $v_t=Q(w_t-w^*)$.

We use $\phi(v)=\frac{\eta C(v)}{2}=\frac{\eta}{2}\Lambda(1+v^T\Gamma v)$, the stationary probability density $p(v)$ satisfies the Smoluchowski equation:
{\small\begin{align}
	0&=\sum_{i=1}^d\frac{\partial}{\partial v_i}\left(\Pi_iv_i\cdot p(v)\right)+\sum_{i=1}^d\frac{\partial}{\partial v_i}\cdot\left(\phi_{i}(w)\frac{\partial}{\partial v_i}p(v)\right) \\
	&=\sum_{i=1}^d\frac{\partial}{\partial v_i}\left(\Pi_{i\cdot}v_i\cdot p(v)\right)+\sum_{i=1}^d\frac{\partial}{\partial v_i}\cdot\left(\frac{\eta\Lambda_i}{2}(1+v^T\Gamma v)\frac{\partial}{\partial v_i}p(v)\right). 
	\end{align}}According to the result for 1-dimensional case, we have the expression of $p(v)$ is  $p(v)\propto(1+v^T\Gamma v)^{-\kappa}$. To determine the value of $\kappa$, we put $p(v)$ in the Smoluchowski equation to obtain
{\small\begin{align*}
	&\sum_{i=1}^d\Pi_{i}p(v)-2\kappa\sum_{i=1}^d\Pi_{i}v_i\cdot \Gamma_{i}v_i\cdot (1+v^T\Gamma v)^{-\kappa-1}\\
	=&\sum_{i=1}^d\frac{\partial}{\partial v_i}\left({\eta\Lambda_{i}\kappa}(1+v^T\Gamma v)^{-\kappa}\cdot \Gamma_{i}v_i\right)\\
	=&\sum_{i=1}^d\left({\eta\Lambda_{i}\kappa}(1+v^T\Gamma v)^{-\kappa}\cdot \Gamma_{i}\right)-2\sum_{i=1}^d\left({\eta\Lambda_{i}\kappa^2}(1+v^T\Gamma v)^{-\kappa-1}\cdot (\Gamma_{i}v_i)^2\right).
	\end{align*}}The we have $\sum_{i=1}^d\Pi_i=\eta\kappa\sum_{i=1}^d\Lambda_i\Gamma_i$. So we have $\kappa=\frac{Tr(H)}{\eta Tr(\Sigma_H\Sigma_g)}$. $\Box$

According to Proposition \ref{prop10},
we can also consider another assumption on $\Sigma_g, \Sigma_H, H$ without assuming their codiagonalization. Instead, we assume (1) If $\Sigma_{ij}=0$, $A^{(ij)}$ is a zero matrix; (2) For $\Sigma_{ij}\neq 0$, $A^{(ij)}$ are equal for all $i\in[d],j\in[d]$ and we denote $A^{(ij)}=\Sigma_H$. We suppose $\eta\cdot \Sigma_H=\kappa H$. (3) $\Sigma_g=\sigma_g\cdot I_d$ which is isotropic. Under these assumptions, we can get the following theorem.
\begin{theorem} (Theorem 4 in main paper)
	If $w$ is $d$-dimensional and  $C(w)$ has the form in Eq.(\ref{eq:c}). The stationary distribution density of multivariate power-law dynamic is 
	\begin{align}
	p(w)=\frac{1}{Z}[1+\frac{1}{\eta\kappa}(w-w^*)^TH\Sigma_g^{-1}(w-w^*)]^{-{\kappa}}  
	\end{align}where $Z=\int_{-\infty}^{\infty}[1+\frac{1}{\eta\kappa}(w-w^*)^TH\Sigma_g^{-1}(w-w^*)]^{-{\kappa}}dw$ is the normalization constant.
\end{theorem}
The proof for Theorem 12 is similar to that for Proposition 11. Readers can check that $p(w)$ satisfies the Smoluchowski equation.


\subsection{Proof for Mean Escaping Time}
\begin{lemma} (Lemma 6 in main paper)
	We suppose {\small$C(w)=\sigma_{g_a}+\frac{2\sigma_{H_a}}{H_a}(L(w)-L(a))$} on the whole escaping path from $a$ to $b$.	The mean escaping time of the 1-dimensional power-law dynamic is,
	{\small\begin{equation}
		\tau=\frac{2\pi}{(1-\frac{1}{2\kappa})\sqrt{{H_a}|H_b|}}\left(1+\frac{2}{\kappa\eta\sigma_{g_a}} \Delta L\right)^{\kappa-\frac{1}{2}},
		\end{equation}}where $\kappa=\frac{H_a}{\eta\sigma_{H_a}}$, $H_a, H_b$ are the second-order derivatives of training loss at local minimum $a$ and saddle point $b$.
\end{lemma}
\textit{Proof:} According to \cite{van1992stochastic}, the mean escaping time $\tau$ is expressed as  {$\tau=\frac{P(w\in V_a)}{\int_{\Omega}J d\Omega}$}, where {$V_a$} is the volume of basin $a$,  {\small$J$} is the probability current that satisfies 
{\small\begin{align*}
	-\nabla J(w,t)&=\frac{\partial}{\partial w}\left(g(w)\cdot p(w,t)\right)+\frac{\partial}{\partial w}\left(\phi(w)\frac{\partial p(w,t)}{\partial w}\right) \\
	&=\frac{\partial}{\partial w}\left(\phi(w)\cdot \left(1+\frac{\mu}{\sigma_g}\Delta L(w)\right)^{-{\kappa}}\frac{\partial\left( \left(1+\frac{\mu}{\sigma_g}\Delta L(w)\right)^{{\kappa}}p(w,t)\right)}{\partial w}\right),
	\end{align*}}where $\phi(w)=\frac{\eta}{2}C(w)$ and $\mu=\frac{2\sigma_{H_a}}{{H_a}}$, $\sigma_g=\sigma_{g_a}$ and $\Delta L(w)=L(w)-L(a)$.  Integrating both sides, we obtain {\small$J(w)=-\phi(w)\cdot \left(1+\frac{\mu}{\sigma_g}\Delta L(w)\right)^{-{\kappa}}\frac{\partial\left( \left(1+\frac{\mu}{\sigma_g}\Delta L(w)\right)^{{\kappa}}p(w,t)\right)}{\partial w}$}.  Because there is no field source on the escape path, $J(w)$ is fixed constant on the escape path. Multiplying $\phi(w)^{-1}\cdot \left(1+\frac{\mu}{\sigma_g}\Delta L(w)\right)^{{\kappa}}$ on both sizes, we have
\begin{align*}
J\cdot \int_{a}^c\phi(w)^{-1}\cdot \left(1+\frac{\mu}{\sigma_g}\Delta L(w)\right)^{{\kappa}}dw&=-\int_a^c \frac{\partial\left( \left(1+\frac{\mu}{\sigma_g}\Delta L(w)\right)^{{\kappa}}p(w,t)\right)}{\partial w}dw\\
&=-0+ p(a).
\end{align*}
Then we get $J=\frac{p(a)}{\int_{a}^c\phi(w)^{-1}\cdot \left(1+\frac{\mu}{\sigma_g}\Delta L(w)\right)^{{\kappa}}dw}$.
As for the term $\int_{a}^c\phi(w)^{-1}\cdot \left(1+\frac{\mu}{\sigma_g}\Delta L(w)\right)^{\frac{1}{\kappa}}dw$, we have 
{\small\begin{align}
	&\int_{a}^c\phi(w)^{-1}\cdot \left(1+\frac{\mu}{\sigma_g}\Delta L(w)\right)^{{\kappa}}dw \label{1}\\
	=&\frac{2}{\eta\sigma_g}\int_{a}^c \left(1+\frac{\mu}{\sigma_g}\Delta L(w)\right)^{-1+{\kappa}}dw \nonumber\\
	\approx&\frac{2}{\eta\sigma_g}\int_{c}^{b} \left(1+\frac{\mu}{\sigma_g}(\Delta L(b)-\frac{1}{2}|H_{b}|(w-b)^2)\right)^{-1+{\kappa}}dw \nonumber\\
	=&\frac{2}{\eta\sigma_g}\int_{c}^{b} \left(1+\frac{\mu}{\sigma_g}(\Delta L(b)-\frac{1}{2}|H_{b}|(w-b)^2)\right)^{-1+{\kappa}}dw  \nonumber\\
	=&\frac{2}{\eta\sigma_g}(1+\frac{\mu}{\sigma_g}\Delta L(b))^{-1+{\kappa}}\int_{c}^{b}\left(1-\frac{\mu}{\sigma_g}\cdot\frac{\frac{1}{2}|H_{b}|(w-b)^2}{1+\frac{\mu}{\sigma_g}\Delta L(b)}\right)^{-1+{\kappa}}dw \nonumber\\
	=&\frac{2}{\eta\sigma_g}(1+\frac{\mu}{\sigma_g}\Delta L(b))^{-1+{\kappa}}\cdot 
	\left(\frac{\frac{1}{2}\frac{\mu}{\sigma_g}|H_{b}|}{1+\frac{\mu}{\sigma_g}\Delta L(b)}\right)^{-1/2}\int_{0}^{1}y^{-1/2}(1-y)^{-1+{\kappa}}dy  \nonumber\\
	=&\frac{2}{\eta\sigma_g}(1+\frac{\mu}{\sigma_g}\Delta L(b))^{-\frac{1}{2}+{\kappa}}\sqrt{\frac{2\sigma_g}{\mu|H_{b}|}}B(\frac{1}{2},{\kappa}), \nonumber
	\end{align}}where the third formula is based on the low temperature assumption. Under the low temperature assumption, we can use the second-order Taylor expansion around the saddle point $b$.

As for the term {\normalsize$P(w\in  V_a)$},  
we have {\small$P(w\in V_a)=\int_{V_a}p(w)dV=\int_{w\in V_a} p(a)(1+\frac{\mu}{\sigma_g}\Delta L(w))^{-\kappa}= p(a)\sqrt{\frac{2\sigma_g}{\mu H_{a}}}B(\frac{1}{2},{\kappa}-\frac{1}{2})$}, where we use Taylor expansion of $L(w)$ near local minimum $a$. 
Then we have $\tau=\frac{P(w\in V_a)}{\int_{\Omega}Jd\Omega}=\frac{P(w\in V_a)}{J}$ because $J$ is a constant. Combining all the results, we can get the result in the lemma.

$\Box$

\begin{theorem} (Theorem 7 in main paper)
	Suppose $w\in\mathbb{R}^d$ and there is only one most possible escaping path between basin $a$ and the outside of basin $a$. The mean escaping time for power-law dynamic escaping from basin $a$ to the outside of basin $a$ is
	{\small\begin{equation}
		\tau=\frac{2\pi\sqrt{-\det(H_b)}}{(1-\frac{d}{2\kappa})\sqrt{\det(H_a)}}\frac{1}{|H_{be}|}\left(1+\frac{1}{\eta\kappa\sigma_{e}} \Delta L\right)^{\kappa-\frac{1}{2}},
		\end{equation}}where $e$ indicates the most possible escape direction, $H_{be}$ is the only negative eigenvalue of $H_b$, $\sigma_{e}$ is the eigenvalue of $\Sigma_{g_a}$ corresponding to the escape direction and $\Delta L=L(b)-L(a)$.
\end{theorem}
\textit{Proof:}
According to \cite{van1992stochastic}, the mean escaping time $\tau$ is expressed as  {$\tau=\frac{P(w\in V_a)}{\int_{\Omega}J d\Omega}$}, where {$V_a$} is the volume of basin $a$,  {\small$J$} is the probability current that satisfies $-\nabla\cdot J(w,t)=\frac{\partial p(w,t)}{\partial t}$.

Under the low temperature assumption, the probability current $J$ concentrates along the direction corresponding the negative eigenvalue of $H_{be}$, and the probability flux of other directions can be ignored. Then we have
\begin{align}
\int_{\Omega}J d\Omega&=J_e\cdot\int_{\Omega}\left(1+\frac{1}{\eta\kappa}(w-b)^T(H_b\Sigma_{g}^{-1})^{\perp e}(w-b)\right)^{-\kappa+\frac{1}{2}}d\Omega,
\end{align}where $J_e=p(a)\cdot \frac{\eta\left(1+\mu\sigma_e\Delta L(b)\right)^{-\kappa+\frac{1}{2}}\sqrt{\mu\sigma_e|H_{be}|}}{2\sqrt{2}B(\frac{1}{2},\kappa)}$ which is obtained by the calculation of $J_e$ for 1-dimensional case in the proof of Lemma 13, and $(\cdot)^{\perp e}$ denotes the directions perpendicular to the escape direction e.

Suppose $H_b\Sigma_g^{-1}$ are symmetric matrix. Then there exist orthogonal matrix $Q$ and diagonal matrix $\Lambda=diag(\lambda_1,\cdots,\lambda_d)$ that satisfy $H_b\Sigma_g^{-1}=Q^T\Lambda Q$. We also denote $v=Q(w-b)$.
We define a sequence as $T_k=1+\frac{1}{\eta\kappa}\cdot\sum_{j=k}^d\lambda_jv_j^2$ for $k=1,\cdots,d$. As for the term $\int_{\Omega}\left(1+\frac{1}{\eta\kappa}(w-b)^T(H_b\Sigma_{g}^{-1})^{\perp e}(w-b)\right)^{-\kappa+\frac{1}{2}}d\Omega$, we have
\begin{align*}
&\int_{\Omega}\left(1+\frac{1}{\eta\kappa}(w-b)^T(H_b\Sigma_{g}^{-1})^{\perp e}(w-b)\right)^{-\kappa+\frac{1}{2}}d\Omega\\
=&\int (1+\frac{1}{\eta\kappa}\cdot v^T\Lambda v)^{-{\kappa}+\frac{1}{2}}dw\\
=&\int (1+\frac{1}{\eta\kappa}\cdot \sum_{j\neq e}^d\lambda_jv_j^2)^{-{\kappa}+\frac{1}{2}}dv\\
=&((\eta\kappa)^{-1}\lambda_1)^{-\frac{1}{2}}\int T_2^{-{\kappa}+\frac{1}{2}}B(\frac{1}{2},{\kappa})dv\\
=&\prod_{j=0}^{d-2}((\eta\kappa)^{-1}\lambda_j)^{-\frac{1}{2}}B(\frac{1}{2},{\kappa}-\frac{j}{2})\\
=&\prod_{j=0}^{d-2}((\eta\kappa)^{-1}\lambda_j)^{-\frac{1}{2}}\cdot\frac{\sqrt{\pi^d}\Gamma({\kappa}-\frac{d}{2})}{\Gamma({\kappa})}\\
=&\frac{\sqrt{(\eta\kappa\pi)^{d-1}}\cdot\Gamma({\kappa}-\frac{d-2}{2})}{\Gamma({\kappa+\frac{1}{2}})\sqrt{\det((H_b\Sigma_g^{-1})^{\perp e})}}.
\end{align*}

As for the term {\normalsize$P(w\in  V_a)$},  
we have 
\begin{align}
&P(w\in V_a)=\int_{V_a}p(w)dV=p(a)\int_{w\in V_a}\left(1+(w-w^*)^TH_a\Sigma_g^{-1}(w-w^*)\right) dw\\
=&p(a)\cdot\frac{\sqrt{(\eta\kappa\pi)^{d}}\cdot\Gamma({\kappa}-\frac{d}{2})}{\Gamma({\kappa})\sqrt{\det((H_a\Sigma_g^{-1}))}}
\end{align}
where we use Taylor expansion of $L(w)$ near local minimum $a$.

Combined the results for $P(w\in V_a)$ and $J$, we can get the result. $\Box$


\subsection{PAC-Bayes Generalization Bound}
We briefly introduce the basic settings for PAC-Bayes generalization error. The expected risk is defined as $\mathbb{E}_{x\sim \mathcal{P}(x)}\ell(w,x)$. Suppose the parameter follows a distribution with density $p(w)$, the expected risk in terms of $p(w)$ is defined as $\mathbb{E}_{w\sim p(w), x\sim \mathcal{P}(x)}\ell(w,x)$. The empirical risk in terms of $p(w)$ is defined as $\mathbb{E}_{w\sim p(w)}L(w)=\mathbb{E}_{w\sim p(w)}\frac{1}{n}\sum_{i=1}^n\ell(w,x_i)$. Suppose the prior distribution over the parameter space is $p'(w)$ and $p(w)$ is the distribution on the parameter space expressing the learned hypothesis function. For power-law dynamic, $p(w)$ is its stationary distribution and we choose $p'(w)$ to be Gaussian distribution with center $w^*$ and covariance matrix $I$. Then we can get the following theorem. 

\begin{theorem} (Theorem 8 in main paper)
	For $w\in\mathbb{R}^d$, we select the prior distribution $p'(w)$ to be standard Gaussian distribution. For $\delta>0$, with probability at least $1-\delta$, the stationary distribution of power-law dynamic has the following generalization error bound,
	\begin{align}\label{eq32}
	\mathbb{E}_{w\sim p(w), x\sim \mathcal{P}(x)}\ell(w,x)\leq\mathbb{E}_{w\sim p(w)}L(w)+\sqrt{\frac{KL(p||p')+\log\frac{1}{\delta}+\log{n}+2}{n-1}},
	\end{align}where $KL(p||p')\leq\frac{1}{2}\log\frac{\det(H)}{\det(\Sigma_g)}+\frac{Tr(\eta\Sigma_gH^{-1})-2d}{4\left(1-\frac{1}{\kappa}\left(\frac{d}{2}-1\right)\right)}+\frac{d}{2}\log\frac{2}{\eta}$ and $\mathcal{P}(x)$ is the underlying distribution of data $x$.
\end{theorem}
\textit{Proof:} Eq.(\ref{eq32}) directly follows the results in \cite{mcallester1999pac}. Here we calculate the Kullback–Leibler (KL) divergence between prior distribution and the stationary distribution of power-law dynamic. The prior distribution is selected to be standard Gaussion distribution with distribution density $p'(w)=\frac{1}{\sqrt{(2\pi)^d\det{(I)}}}\exp\{-\frac{1}{2}(w-w^*)^TI(w-w^*)\}$. The posterior distribution density is the stationary distribution for power-law dynamic, i.e., $p(w)=\frac{1}{Z}\cdot(1+\frac{1}{\eta\kappa}\cdot (w-w^*)^TH\Sigma_g^{-1} (w-w^*))^{-{\kappa}}$. 

Suppose $H\Sigma_g^{-1}$ are symmetric matrix. Then there exist orthogonal matrix $Q$ and diagonal matrix $\Lambda=diag(\lambda_1,\cdots,\lambda_d)$ that satisfy $H\Sigma_g^{-1}=Q^T\Lambda Q$. We also denote $v=Q(w-w^*)$.

We have
\begin{align*}
&\log\left(\frac{p(w)}{p'(w)}\right)\\&=-\kappa\log(1+\frac{1}{\eta\kappa}\cdot (w-w^*)^TH\Sigma_g^{-1}(w-w^*))-\log Z+\frac{1}{2}(w-w^*)^TI(w-w^*)+\frac{d}{2}\log 2\pi 
\end{align*}

The KL-divergence is defined as $KL(p(w)||p'(w))=\int_wp(w)\log \left(\frac{p(w)}{p'(w)}\right)dw$. Putting $v=Q(w-w^*)$ in the integral, we have
\begin{align}
&KL(p(w)||p'(w)) \nonumber\\
=&\frac{d}{2}\log 2\pi-\log Z+\frac{1}{2Z}\int_vv^Tv\left(1+\frac{1}{\eta\kappa}\cdot v^T\Lambda v\right)^{-{\kappa}}dv-\frac{1}{Z\eta}\int_vv^T\Lambda v\cdot(1+\frac{1}{\eta\kappa}\cdot v^T\Lambda v)^{-\kappa}dv, \label{eq:KL}
\end{align}where we use the approximation that $\log(1+x)\approx x$.
We define a sequence as $T_k=1+\frac{1}{\eta\kappa}\cdot\sum_{j=k}^d\lambda_jv_j^2$ for $k=1,\cdots,d$. We first calculate the normalization constant $Z$.
\begin{align*}
Z=&\int (1+\frac{1}{\eta\kappa}\cdot v^T\Lambda v)^{-{\kappa}}dw=\int (1+\frac{1}{\eta\kappa}\cdot \sum_{j=1}^d\lambda_jv_j^2)^{-{\kappa}}dv\\
=&((\eta\kappa)^{-1}\lambda_1)^{-\frac{1}{2}}\int T_2^{-{\kappa}+\frac{1}{2}}B(\frac{1}{2},{\kappa}-\frac{1}{2})dv=\prod_{j=1}^d((\eta\kappa)^{-1}\lambda_j)^{-\frac{1}{2}}B(\frac{1}{2},{\kappa}-\frac{j}{2})\\
=&\prod_{j=1}^d((\eta\kappa)^{-1}\lambda_j)^{-\frac{1}{2}}\cdot\frac{\sqrt{\pi^d}\Gamma({\kappa}-\frac{d}{2})}{\Gamma({\kappa})}
\end{align*}

We define $Z_j=((\eta\kappa)^{-1}\lambda_j)^{-\frac{1}{2}}B\left(\frac{1}{2},{\kappa}-\frac{j}{2}\right)$.
For the third term in Eq.(\ref{eq:KL}), we have
{\small\begin{align*}
	&2Z\cdot\uppercase\expandafter{\romannumeral 3}\\
	=&\int_vv^Tv(1+\frac{1}{\eta\kappa} v^T\Lambda v)^{-{\kappa}}dv\\
	=&\int_{v_2,\cdots v_d}\int_{v_1}v_1^2\left(1+\frac{1}{\eta\kappa}\cdot v^T\Lambda v\right)^{-{\kappa}}dv_1+Z_1\left(\sum_{j=2}^dv_j^2\right)\left(1+\frac{1}{\eta\kappa}\cdot\sum_{j=2}^d\lambda_jv_j^2\right)^{-{\kappa}+\frac{1}{2}}d_{v_2\cdots,v_d}\\
	=&\int_{v_2,\cdots v_d}T_2^{-{\kappa}}\int_{v_1}v_1^2\left(1+\frac{(\eta\kappa)^{-1}\lambda_1v_1^2}{T_2}\right)^{-{\kappa}}dv_1+Z_1\left(\sum_{j=2}^dv_j^2\right)\left(1+\frac{1}{\eta\kappa}\cdot\sum_{j=2}^d\lambda_jv_j^2\right)^{-{\kappa}+\frac{1}{2}}d_{v_2\cdots,v_d}\\
	=&\int_{v_2,\cdots,v_d}T_2^{-{\kappa}}\int \left(\frac{T_2}{(\eta\kappa)^{-1}\lambda_1}\right)^{\frac{3}{2}}y^\frac{1}{2}\left(1+y\right)^{-{\kappa}}dy+Z_1\left(\sum_{j=2}^dv_j^2\right)\left(1+\frac{1}{\eta\kappa}\cdot\sum_{j=2}^d\lambda_jv_j^2\right)^{-{\kappa}+\frac{1}{2}}d_{v_2\cdots,v_d}\\
	=&\int_{v_2,\cdots,v_d}((\eta\kappa)^{-1}\lambda_1)^{-\frac{3}{2}}T_2^{-{\kappa}+\frac{3}{2}}B\left(\frac{3}{2},{\kappa}-\frac{3}{2}\right)+Z_1\left(\sum_{j=2}^dv_j^2\right)\left(1+\frac{1}{\eta\kappa}\cdot\sum_{j=2}^d\lambda_jv_j^2\right)^{-{\kappa}+\frac{1}{2}}d_{v_2\cdots,v_d}\\
	=&(\frac{\lambda_1}{\eta\kappa})^{-\frac{3}{2}}B\left(\frac{3}{2},{\kappa}-\frac{3}{2}\right)\int_{v_2,\cdots,v_d}T_2^{-{\kappa}+\frac{3}{2}}d_{v_2\cdots,v_d}+\int_{v_2,\cdots,v_d}Z_1\left(\sum_{j=2}^dv_j^2\right)\left(1+\frac{1}{\eta\kappa}\cdot\sum_{j=2}^d\lambda_jv_j^2\right)^{-{\kappa}+\frac{1}{2}}d_{v_2\cdots,v_d}
	\end{align*}} 
For term $\int_{v_2,\cdots,v_d}T_2^{-\frac{1}{\kappa}+\frac{3}{2}}d_{v_2\cdots,v_d}$ in above equation, we have 
\begin{align*}
&\int_{v_2,\cdots,v_d}T_2^{-{\kappa}+\frac{3}{2}}d_{v_2\cdots,v_d}\\
=&\int_{v_3,\cdots,v_d}T_3^{-{\kappa}+2}((\eta\kappa)^{-1}\lambda_2)^{-\frac{1}{2}}B\left(\frac{1}{2},{\kappa}-2\right)d_{v_3,\cdots,v_d}\\
=&\int_{v_4,\cdots,v_d}T_4^{-{\kappa}+\frac{5}{2}}((\eta\kappa)^{-1}\lambda_2)^{-\frac{1}{2}}((\eta\kappa)^{-1}\lambda_3)^{-\frac{1}{2}}B\left(\frac{1}{2},{\kappa}-\frac{5}{2}\right)B\left(\frac{1}{2},{\kappa}-2\right)d_{v_4,\cdots,v_d}\\
=&\int_{v_d}T_d^{-{\kappa}+\frac{1}{2}+\frac{1}{2}\times d}\prod_{j=2}^{d-1}((\eta\kappa)^{-1}\lambda_j)^{-\frac{1}{2}}\prod_{j=2}^{d-1}B\left(\frac{1}{2},{\kappa}-(\frac{j}{2}+1)\right)d_{v_d}\\
=&\prod_{j=2}^{d}((\eta\kappa)^{-1}\lambda_j)^{-\frac{1}{2}}\prod_{j=2}^{d}B\left(\frac{1}{2},{\kappa}-(\frac{j}{2}+1)\right)
\end{align*}
Let $A_j=((\eta\kappa)^{-1}\lambda_j)^{-\frac{3}{2}}B\left(\frac{3}{2},{\kappa}-(\frac{j}{2}+1)\right)$. According to the above two equations, we can get the recursion
\begin{align*}
&2Z\int v^TvT_1^{-{\kappa}}dv\\=&A_1\cdot\int T_2^{-{\kappa}+\frac{3}{2}}+Z_1\int_{v_2,\cdots,v_d}\left(\sum_{j=2}^dv_j^2\right)T_2^{-{\kappa}+\frac{1}{2}}d_{v_2\cdots,v_d}\\
=&A_1\cdot\int T_2^{-{\kappa}+\frac{3-1}{2}}d_{v_2\cdots v_d}+Z_1\cdot A_2\int T_3^{-{\kappa}+\frac{4}{2}}d_{v_3\cdots,v_d}+Z_1Z_2\int\left(\sum_{j=3}^dv_j^2\right)T_3^{-{\kappa}+\frac{1}{2}}d_{v_3\cdots,v_d}\\
=&\sum_{j=1}^{d-1} A_j\prod_{k=1}^{j-1}Z_k\int T_{j+1}^{-{\kappa}+\frac{j+1+1}{2}}d_{v_{j+1},\cdots,v_d}+\prod_{k=1}^{d-1}Z_k\int v_d^2T_d^{-{\kappa}+\frac{d-1}{2}}dv_d\\
=&\sum_{j=1}^{d-1} (\frac{\lambda_j}{\eta\kappa})^{-\frac{3}{2}}B\left(\frac{3}{2},{\kappa}-(\frac{j}{2}+1)\right)\prod_{k=1}^{j-1}(\frac{\lambda_k}{\eta\kappa})^{-\frac{1}{2}}B\left(\frac{1}{2},{\kappa}-\frac{k}{2}\right)\prod_{s=j+1}^{d}((\frac{\lambda_s}{\eta\kappa})^{-\frac{1}{2}}\prod_{s=j+1}^{d}B\left(\frac{1}{2},{\kappa}-(\frac{s}{2}+1)\right)\\
+&\prod_{j=1}^{d-1}(\frac{\lambda_j}{\eta\kappa})^{-\frac{1}{2}}B(\frac{1}{2},{\kappa}-\frac{j}{2}-1)\cdot (\frac{\lambda_d}{\eta\kappa})^{-\frac{3}{2}}B(\frac{3}{2},{\kappa}-(\frac{d}{2}+1))\\
=&\frac{\sqrt{\pi^{d}}\Gamma({\kappa}-\frac{d}{2}-1)Tr(H^{-1}\Sigma_g)}{2\Gamma({\kappa})\sqrt{(\eta\kappa)^{-(d+2)}\det(H^{-1}\Sigma_g)}}
\end{align*}
We have
\begin{align*}
III=&\frac{\sqrt{\pi^{d}}\Gamma({\kappa}-\frac{d}{2}-1)Tr(H^{-1}\Sigma_g)}{4\Gamma({\kappa})\sqrt{(\eta\kappa)^{-(d+2)}\det(H^{-1}\Sigma_g)}} \cdot \prod_{j=1}^d((\eta\kappa)^{-1}\lambda_j)^{\frac{1}{2}}\cdot\frac{\Gamma({\kappa})}{\sqrt{\pi^d}\Gamma({\kappa}-\frac{d}{2})}\\
=&\frac{\eta\kappa Tr(H^{-1}\Sigma_g)}{4({\kappa}-\frac{d}{2}-1)}
\end{align*}
Similarly, for the fourth term in Eq.(\ref{eq:KL}), we have $IV=\frac{\kappa d}{2(\kappa-\frac{d}{2}-1)}$. Combining all the results together, we can get $KL(p||p')=\frac{1}{2}\log\frac{\det(H)}{(\eta\kappa)^d\det(\Sigma_g)}+\log\frac{\Gamma(\kappa)}{\Gamma(\kappa-\frac{d}{2})}+\frac{Tr(\eta\Sigma_gH^{-1})-2d}{4\left(1-\frac{1}{\kappa}\left(\frac{d}{2}-1\right)\right)}+\frac{d}{2}\log2$. Using the fact that $\log\frac{\Gamma(\kappa)}{\Gamma(\kappa-\frac{d}{2})}\leq\frac{d}{2}\log\kappa$, we have $KL(p||p')\leq\frac{1}{2}\log\frac{\det(H)}{\det(\Sigma_g)}+\frac{Tr(\eta\Sigma_gH^{-1})-2d}{4\left(1-\frac{1}{\kappa}\left(\frac{d}{2}-1\right)\right)}+\frac{d}{2}\log\frac{2}{\eta}$.

\subsection{Implementation Details of the Experiments}
\subsubsection{Observations on the Covariance Matrix}
In this section, we introduce the settings on experiments of the quadratic approximation of covariance of the stochastic gradient on plain convolutional neural network (CNN) and ResNet. For each model, we use gradient descent with small constant learning rate to train the network till it converges. The converged point can be regarded as a local minimum, denoted as $w^*$.

As for the detailed settings of the CNN model, the structure for plain CNN model is $input\rightarrow Conv1\rightarrow maxpool\rightarrow Conv2\rightarrow maxpool\rightarrow fc1\rightarrow Relu\rightarrow fc2\rightarrow output$. Both $Conv1$ and $Conv2$ use $5\times 5$ kernels with 10 channels and no padding. Dimensions of full connected layer $fc1$ and $fc2$ are $1600\times 50$ and $50\times 10$ respectively. We randomly sample 1000 images from FashionMNIST \cite{xiao2017fashion} dataset as training set. The initialization method is the Kaiming initialization \cite{he2015delving} in PyTorch. The learning rate of gradient descent is set to be $0.1$. After $3000$ iterations, GD converges with almost $100\%$ training accuracy and the training loss being $1e^{-3}$.

As for ResNet, we use the ResNet-18 model \cite{he2016deep} and randomly sample 1000 images from Kaggle's dogs-vs-cats dataset as training set.  The initialization method is the Kaiming initialization \cite{he2015delving} in PyTorch. The learning rate of gradient descent is set to be $0.001$. After $10000$ iterations, GD converges with $100\%$ training accuracy and the training loss being $1e^{-3}$.	

We then calculate the covariance matrix of the stochastic gradient at some points belonging to the local region around $w^*$. The points are selected according to the formula:  $w^*_{layerL}\pm (i\times Scale)$, where $w^*_{layerL}$ denotes the parameters at layer $L$, and $i\times Scale, i\in[N]$ determines the distance away from $w_{layerL}^*$. When we select points according to this formula by changing the parameters at layer $L$, we fixed the parameters at other layers. For both CNN model and ResNet18 model, we select $20$ points by setting $i=1,\cdots,10$. For example, for CNN model, we choose the 20 points by changing the parameters at the $Conv1$ layer with $Scale=0.001$ and $Conv2$ layer with $Scale=0.0001$, respectively. For ResNet18, we choose the 20 points by changing the parameters for a convolutional layer at the first residual block with $Scale=0.0001$ and second residual block with $Scale=0.0001$, respectively.



The results are shown in Figure.\ref{fig:2dmodel}. The x-axis denotes the distance of the point away from the local minimum and the y-axis shows the value of the trace of covariance matrix at each point. The results show that the covariance of noise in SGD is indeed not constant and it can be well approximated by quadratic function of state (the blue line in the figures), which is consistent with our theoretical results in Section 3.1.

\subsubsection{Supplementary Experiments on Parameter Distributions of Deep Neural Networks}
\begin{figure}[h]
	\centering
	\includegraphics[width=0.9\textwidth]{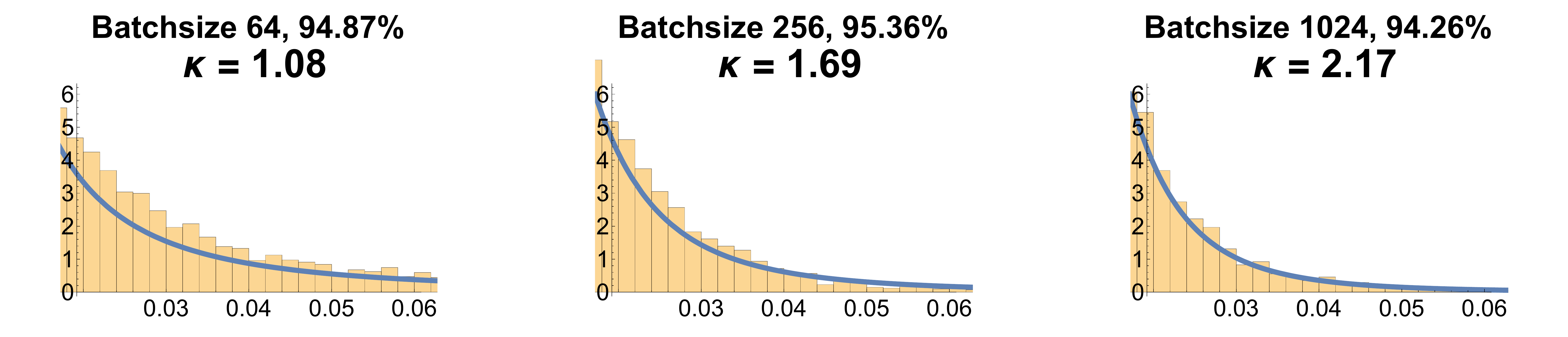}
	\caption{
		A close-up of right tail distribution of the result for ResNet18 in Figure. \ref{fig:pl1}, which could help to observe the heavy-tailed properties among different batchsize.}
	\label{fig:pltail}
\end{figure}
\begin{figure}[h]
	\centering
	\includegraphics[width=0.9\textwidth]{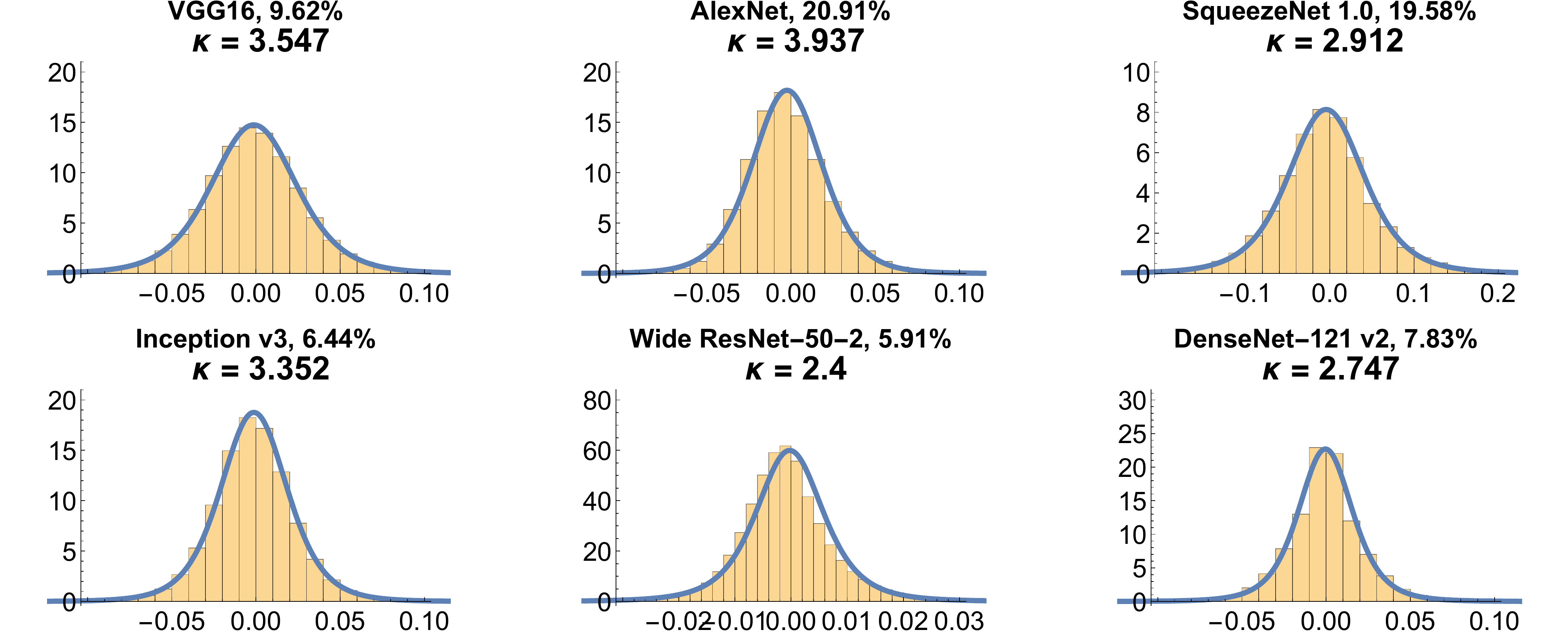}
	\caption{Approximating distribution of parameters (trained by SGD) by power-law dynamic. These networks use pre-trained models offered by PyTorch, and all of them are pre-trained on ImageNet dataset using SGD. Percentage on the right side of network name is Top-5 test error of each network. }
	\label{fig:plpretrain}
\end{figure}
For Figure. \ref{fig:pl1}, we train LeNet-5 on MNIST dataset using SGD with constant learning rate $\eta = 0.03$ for each batchsize till it converges. Parameters are $conv2.weight$ in LeNet-5. For Figure \ref{fig:pl2}, we train ResNet-18 on CIFAR10 using SGD with momentum. We do a $RandomCrop$ on training set scaling to $32\times 32$ with $padding = 4$ and then a $RandomHorizontalFlip$. In training, momentum is set to be $0.9$ and weight decay is set to be $5e-4$. Initial learning rate in SGD is set to be $0.1$ and we using a learning rate decay of $0.1$ on $\{150, 250\}$-th epoch respectively. We train it until converges after 250 epoch. Parameters are $layer1.1.conv2.weight$ in ResNet-18.

We also observe the parameter distribution on many pretrained models. Details for pre-trained models can be found on \url{https://pytorch.org/docs/stable/torchvision/models.html}. Figure.\ref{fig:plpretrain} shows the distribution of parameters trained by SGD can be well fitted by power-law distribution.  Parameters in this figure are all randomly selected to be \textit{features.10.weight}, $\textit{features.14.weight}$, $features.5.expand3\times 3.weight$, $Mixed\_6d.branch7\times 7\_3.conv.weight$, $layer4.2.conv3.weight$  and $features.denseblock2.denselayer1.conv2.weight$ for VGG-16, AlexNet, SqueezeNet 1.0, Inception v3, Wide ResNet-50-2 and DenseNet-121 respectively.

\subsubsection{Further Explanation on Experiments in Section 5.2}
As for the experiments for 2-D model, we also calculate coefficient of the second-order term for the quadratic curve shown in Figure.\ref{fig:cov1}, and its value is roughly $30$, which matches the result in Figure.\ref{fig:succn1} in the sense that the result for SGD is similar with the result for power-law dynamic with $\lambda_1\approx 32$.

\end{document}